%% file: DeepAndGPBA.tex
\documentclass[journal]{IEEEtran}

\usepackage{makecell}
\usepackage[hidelinks]{hyperref}
\usepackage{comment}

\usepackage{fixltx2e}
\usepackage{graphicx}
\usepackage{caption}
\usepackage{subcaption}
\usepackage{mathtools}
\DeclarePairedDelimiter\ceil{\lceil}{\rceil}

\IEEEoverridecommandlockouts
\usepackage{amssymb,amsmath,amsthm}
\usepackage[font=small,labelfont=bf]{caption}

\newtheorem{ppro}{Proposition}

\usepackage{soul}
\makeatletter
\newcommand{\biggg}[1]{{\hbox{$\left#1\vbox to 20.5pt{}\right.\n@space$}}}
\newcommand{\Biggg}[1]{{\hbox{$\left#1\vbox to 23.5pt{}\right.\n@space$}}}
\newcommand{\bigggg}[1]{{\hbox{$\left#1\vbox to 26.5pt{}\right.\n@space$}}}
\newcommand{\Bigggg}[1]{{\hbox{$\left#1\vbox to 29.5pt{}\right.\n@space$}}}
\newcommand{\biggggg}[1]{{\hbox{$\left#1\vbox to 32.5pt{}\right.\n@space$}}}
\newcommand{\Biggggg}[1]{{\hbox{$\left#1\vbox to 35.5pt{}\right.\n@space$}}}
\newcommand{\bigggggg}[1]{{\hbox{$\left#1\vbox to 38.5pt{}\right.\n@space$}}}
\newcommand{\Bigggggg}[1]{{\hbox{$\left#1\vbox to 41.5pt{}\right.\n@space$}}}
\makeatother

\usepackage{bm,cite,algorithm,algorithmic,float,amsmath,amssymb}
\usepackage{mdwmath}
\usepackage{mdwtab}
\usepackage{xcolor}
\usepackage{cool}
\usepackage{makeidx,enumitem}
\usepackage{balance}
\floatname{algorithm}{Algorithm}

\makeatletter
\renewcommand\paragraph{\@startsection{paragraph}{4}{\z@}%
            {-2.5ex\@plus -1ex \@minus -.25ex}%
            {1.25ex \@plus .25ex}%
            {\normalfont\normalsize\itshape}}
\makeatother
\usepackage{multirow}
\usepackage{epstopdf}

\makeatletter
\newcommand\notsotiny{\@setfontsize\notsotiny{7}{6.5}}
\makeatother
\begin{document}
\title{Supervised ML Solution for Band Assignment in Dual-Band Systems with Omnidirectional and Directional Antennas  
 \thanks{D. Burghal,  R. Wang,  A. F. Molisch are at the Ming Hsieh Department of Electrical Engineering and Computer, University of Southern California, Los Angeles, CA 90089, USA. (\{burghal,wang78,molisch\}@usc.edu).}
 
 \thanks{ A. Alghafis is with the Communication and Information Technology Research Institute,  KACST, Riyadh 11442, KSA (alghafis@kacst.edu.sa).}}

\author{
\IEEEauthorblockN{Daoud Burghal,  Rui Wang,  Abdullah  Alghafis  and Andreas F. Molisch \emph{Fellow, IEEE}}
}
\maketitle
\vspace{-10mm}
\begin{abstract}
Many wireless networks, including 5G NR (New Radio) and future beyond 5G cellular systems, are expected to operate on multiple frequency bands. This paper considers the band assignment (BA) problem in dual-band systems, where the basestation (BS) chooses one of the two available frequency bands (centimeter-wave and millimeter-wave bands) to communicate with the user equipment (UE). While the millimeter-wave band might offer higher data rate, there is a significant probability of outage during which the communication should be carried on the (more reliable) centimeter-wave band. With mobility, the BA can be perceived as a sequential problem, where the BS uses previously observed information to predict the best band for a future time step. 
 
We formulate the BA as a binary classification problem and propose supervised Machine Learning (ML) solutions. We study the problem when both the BS and the UE use (i) omnidirectional antennas and (ii) both use directional antennas. In the omnidirectional case, we derive analytical benchmark solutions based on the Gaussian Process (GP) assumption for the inter-band shadow fading. In the directional case, where the labeling is shown to be complex, we propose an efficient labeling approach based on the Viterbi Algorithm (VA). We compare the performances for two channel models: (i) a stochastic channel and (ii) a ray-tracing based channel.  
\end{abstract}
\vspace{-0mm}

\vspace{-5mm}
\section{Introduction}
\input{nIntroduction}

\subsection{Prior Work}
\input{nPriorWork}

\subsection{Contribution and Paper Structure}
\input{nContribution}

\section{Problem and Solutions Overview}\label{sec:overview}
\subsection{Basic System Model}
\input{BasicSystemModel}

\subsection{Problem Description}
\input{ProblemDesc}

\subsection{Performance Metrics}\label{subsec:DLPerfMet}
\input{PerformanceMetric}

\subsection{Solutions Overview}
\input{SolutionOverview}
 
\section{Data Aspects}\label{sec:DataAspects}
In this section we cover several aspects of the data. We first highlight the different environments and features, then we discuss the proposed methods to create the labels and the sequences.
\subsection{Environments and Features}
We consider two environments, the first is a stochastic environment that we use to compare the ML solution to the GP-based benchmark for the SISO system, the second is a ray-tracing based solution that we use for both SISO and MIMO systems.

\subsubsection{Stochastic Based Environment}\label{sec:stochData}
In this synthetic environment we generate the data such that it matches the GP assumptions, which can be a reasonable assumption in some scenarios \cite{burghal2019dual}.
\input{Stoch_TheEnv}
The above generated shadowing realizations are jointly normal over the two bands. However, due to spatial filtering and the selection of the best beam, these assumptions are not valid for MIMO systems, thus in Sec. \ref{Sec:StochEnvDL} we limit the discussion to SISO systems. We consider four features (input to the learning solutions): ($\rm f_1$) the distance from the BS to the UE $d$ in meters, ($\rm f_2$) the angular position of the UE $\theta$ in rad, ($\rm f_3$) the received signal strength (or the SNR) in the cmWave band in dBm (or dB), and ($\rm f_4$) is the received signal strength (or the SNR) in the mmWave band in dBm (or dB). 

\subsubsection{Ray-tracing Based Environment}
\input{RT_TheEnv}

In Sec. \ref{Sec:RTEnvDL} we use this environment for SISO and MIMO systems. In addition to the features ($\rm f_1$) to ($\rm f_4$) above, we here use ($\rm f_5$) the delay of the dominant MPC in seconds, and ($\rm f_6$) the AoA of the dominant MPC, where the dominant MPC is the one with the highest power. 

\subsubsection{Feature Availability and Pre-processing}\label{sec:featF}
The availability of the features depends on the system implementations. For instance, ($\rm f_1$) and ($\rm f_2$), i.e., $(d,\theta)$ (which represent the polar coordinates of the UE with respect to the BS), may be estimated using signal processing techniques or acquired by explicit feedback of the GPS data. To extract ($\rm f_5$) large bandwidth might be required, for ($\rm f_6$) the use of antenna arrays is necessary. Using both ($\rm f_3$) and ($\rm f_4$) may require additional effort or equipment at the UE side. We consider several combinations of the above features and discuss their effectiveness for BA.

As typically done in ML, we perform pre-processing of the features, in particular we standardize the input features such that their mean is zero and the standard deviation is equal to one. In addition, we utilize the prior knowledge about the wireless propagation, for instance we use logarithmic scale for distances and power, as this may linearize their relation with one another.

\subsection{Generating Sequences}\label{sec:genSeq}
\input{Trajectories}


\subsection{Labeling}\label{sec:labeling}
Labels are key components of supervised ML solutions. They should capture the essential system model, they also impact the solution quality and the overall performance. For clarity and based on the discussion in Sec. \ref{sec:overview} we distinguish two cases.
\subsubsection{SISO System}
In this case, for simplicity, we neglect the switching cost. As a result labeling is straightforward: at $t$ the system will assume $\mathcal{L}_{{\rm s} j}(t) = 1$ when $R^m(t+U) \geq R^c(t+U)$.\footnote{We here emphasize that the prediction at time frame $t$ is for time frame $t+U$.}

\subsubsection{MIMO System}
For a realistic scenario with antenna arrays in both frequency bands and switching cost, our goal is to maximize the sum rate of the UE by the end of the session (a mobility sequence). To quantify the beamforming and switching cost, we propose a model based on the 3GPP NR standard \cite{dahlman20205g}, where we assume that the BS will frequently transmit beamformed reference signals (e.g., Synchronization Signal Block (SSB) \cite{dahlman20205g}) that the UE has to monitor to assess the best beam that it should use. In that case, the achievable rate in time $t$ in band $b$ is 
\begin{align}\label{eq:ratediscount} 
\tilde{ {R}^b_j}(t) =   {R}^b_j(t)   \frac{T_f- T_s^b}{T_f}, 
\end{align}
where $T_f$ is the duration of the time-frame and $T_s^b$ is the beam alignment cost in band $b$, which depends on the number of directions (beams)/antennas and the search technique. In practice, full beam sweep ("major" beam search) is not always necessary as it is possible to do a "minor" search within a candidate list of beam combinations; this is especially necessary for large number of antennas. Let $\mathcal{C}^b(t)$ be the list of candidate beams in band $b$ at time $t$, furthermore, let $T_{\rm s}^b = T^b_{\rm M}$ and $T_{\rm s}^b = T^b_{\rm m}$ represent the time needed for major and minor beam search, respectively. The list $\mathcal{C}^b(t)$ may include the past successful beams and their neighbors.

For a given sequence of features and rates, we need to find a good labeling technique. This is challenging as the decision/label at $t$ impacts (i) the possible candidate list for future time instances, (ii) the set of observable features, and (iii) the overall switching decisions as. For example, it is reasonable to stay on one band if the UE needs to be switched back to it later, i.e., to minimize the frequent switching cost. One solution is to list all possible switching decisions and their relative costs to create the labeled sequences. However, this brute-force solution to the combinatorial problem is not feasible for reasonable sequence lengths. Here we make a set of simplification assumptions and propose an {\em efficient} labeling solution.

While practically deployed methods to handle the required frequent beam search are usually proprietary to the service provider/vendors and depend on system design, we here propose a method that in essence only depends on basic NR assumption, where we utilize the periodic SSB to schedule frequent beam search (or a periodic CSI-RS). Thus, the system performs a periodic major search every $N^b_{\rm p}$ frames in band $b$; in between the UE and BS do a minor search in $\mathcal{C}^b$ every frame. Major searches could also be performed if the received signal power is below a certain threshold (e.g., detectable signal level), i.e., link failure, or when the UE switches to a new band. Note that both the size of $\mathcal{C}^b$ and values of $T_s^b$ depend on the number of antennas/beams. We here set the size $\mathcal{C}^b$ such that the device experiences an outage with a probability of less than $Z$ (e.g., outage with probability $\leq$ $0.1$). Fig. \ref{fig:mnrMajr} illustrates the idea of periodic beam search. More implementation details are presented in Sec. \ref{subsec:ImpMIMOdetails}.

With the above assumptions in mind, we utilize the VA to perform the sequence search and label the sequences. We consider four states that represent the major and minor searches in cmWave and mmWave, see Fig. \ref{fig:VAstates}. The edges that connect the states represent the transition between them over time. Whenever the UE switches between the bands, a major search is triggered in the new band. The cost could also be made to include additional delay due to the physical layer (e.g., synchronization) or upper layer (e.g., protocol procedures); nevertheless, the labeling process is the same, but the value of $T^b_M$ should be modified to include additional delays for that particular edge. In addition to the periodic major search and band switching, major search in the band is also triggered in case of an outage (the power level of all the beams in the available candidate list drops below a certain threshold). We emphasize here that although the VA has four states, the label could have one of two values $\{0,1\}$, which is determined based on the operating band after the transitions. In appendix-C, we provide a worked-out labeling example for demonstration.

\input{LabelingDiagrams}

Once the states are identified, we select the decisions (labels) that maximize the sum rate for a given sequence. Finally, we point out that the VA is a {\em heuristic} solution for this problem, as it does not take into account the impact of the future decisions during the selection process of the paths at each state. However, VA provides a sequential decision tool that can be used to identify the sequences and incorporate the rates and losses. Other solutions, including algorithms that consider forward and backward passes, are interesting future study directions.

\section{Machine Learning Solution}\label{sec:MLsolution}
Before we dive into the details of the proposed ML architecture, we first briefly introduce of the used ML algorithms and the datasets notations.

\subsection{Preliminaries}

\input{Preliminaries}

\section{SISO Benchmark: Gaussian Process Based BA}\label{sec:GPBasedSol}
\input{GP_Based_BA}

\section{Experiment I: Stochastic Environment}\label{Sec:StochEnvDL}
Here we study the performance of the solutions using stochastically generated channels. This will provide a comparison between the learning based and the GP based solutions in a synthetic environment that matches the GP assumptions. We first describe the general data generation model and then address the dataset details and the solutions performance. For the learning schemes, we emphasize that, the displayed performance values are not guaranteed to be optimal, as we have considered a limited number of structures and parameters.

\input{Stoch_Seq}

 \section{Experiment II: Simulated Campus Environment}\label{Sec:RTEnvDL}

%
\subsection{SISO System}\label{sec:SISO}
\input{RT_Seq}

\subsection{MIMO System}
\subsubsection{Data Generation Details} \label{subsec:ImpMIMOdetails}
We use the same methods to generate the sequences as above, however, we assume that the system uses $20$ ms SSB periodicity during which the BS can schedule up to $64$ beams in the mmWave and up to four beams in the cmWave band, respectively. To sweep beams at the UE side or in case of more beams at the BS side, multiples of this period are needed. Thus the cost of major search in band $b$ is
\begin{align}\label{eq:TbM}
    T^b_{\rm M} = \ceil*{\frac{ {K}^b }{B^b}}  {M}^b, 
\end{align}
where $B^m $= 64 and $B^c$= 4. For a minor search we assume that the size of the $\mathcal{C}^b$ is $0.125$ of the total beam combinations, i.e., $T^b_{\rm m} = \frac{1}{8} T^b_{\rm M}$. The beams in $\mathcal{C}^b$ are chosen based on the best beams in a previous full sweep, and a subset of their neighbors. The beams are assumed to be ideal with zero gain outside the main-lobes, beamwidths equal to $\frac{2\pi}{{K}^b } $ for the BS and $\frac{2\pi}{ {M}^b } $ for the UE.

We declare the link to be in an outage when the received power falls below $-120- 10\log_{10}( {K}^b {M}^b )$ dBm. In that case a full beam sweep in band $b$ is used. In the following we focus on a system with $K^m\times M^m : 128\times 32$ and $K^c\times M^c : 16\times 4$.\footnote{We here deviate from the typical initial access beams that are restricted to $20$ ms for full BS sweep.} The periodicity of the full beam sweep is chosen at $4$ and $6$ time frames in mmWave and cmWave bands, respectively, where the simulation shows that the outage probability is less than $0.1$, i.e., $Z = 0.1$ (see Sec. \ref{sec:labeling}).

Other values of the above parameters could be used with no or minimal system modifications. For the purpose of comparability, we now assume that the transmitted power values are reduced by the beam gain, which can be chosen such that the maximum received SNRs are the same as in Sec. \ref{sec:SISO}, i.e., for mmWave it is reduced by $36$ dB, and for cmWave by $18$ dB. Nevertheless, although this will {\em roughly} maintain the ratio of mmWave to cmWave labels in the grid, it is still difficult to compare the performance as functions of array sizes (including SISO) since: (i) the size of the candidate list and thus the outage probability depends on the array size, and (ii) the labels and the sequences depend on the beamwidth, which impacts the number of MPCs per beam and thus the achievable rates.

\subsubsection{Performance}
The performance of the ML solutions using different feature combinations is presented in Table \ref{Tab:MIMOSeqResultsRT}. The table shows the mis-classification error and the normalized rate loss at $U = 4$. Before discussing the details of the ML solutions, we consider the following extreme benchmarks. The normalized rate loss for all zero decisions (i.e., using cmWave only) and all ones decisions (i.e., using mmWave only) are $0.28$ and $0.22$, respectively, where the loss is with respect to the rates obtained using the labeling method of Sec. \ref{sec:labeling}. Another interesting comparison is between the accumulated rates with and without the cost values (i.e., setting $T^b_{\rm M}$ and $T^b_{\rm m}$ to zero). The normalized rate loss, with respect to the latter (loss free), is about $0.12$, indicating degradation due to the frequent beam search and switching cost.

\input{Tables/TblMIMOSeq}

ML solutions demonstrate the advantage of dual band systems with a rate loss that is (in the worst case) better than for a single band system, despite the increasing switching cost. Compared to SISO, the mis-classification error has increased in general. Similar to SISO, we notice that the combinations that have the location and the omni-directional power information or its equivalent (delay and AoA) are showing better results. However, note that with MIMO systems, assuming the knowledge of AoA in the cmWave band might be difficult if a small array is used there, making combinations (c-$7$) to (c-$9$) potentially impractical. However, even in this case, we can achieve good BA prediction with alternative features combinations.

Regarding the underlying ML method, the LSTM solution is able to achieve better than $0.08$ rate loss with almost all feature combinations, something that is not feasible with simpler architectures, showing, again, the applicability of LSTM based method to this complex sequential prediction problem. This is especially with simple feature combinations. However, we notice that the relative performance of the different solutions is maintained, the NN method showing comparable performance with certain features and the GR method suffering from large losses. 

\section{Conclusion}\label{Sec:ConcDL}
{ In this paper, we propose a supervised ML solution for the sequential BA in dual-band omni-directional (SISO) and directional (MIMO) systems. We formulated the BA as a sequential binary classification problem and proposed an ML structure that uses LSTM network. To enable the supervised solution, we propose a system model based on the basic 5G NR structure and put forth a labeling method to label the sequences while considering the switching cost, e.g., due to the frequent beam-search in a MIMO system. We study the performance of the solution against a number of benchmarks and with different feature combinations. For SISO, we derive an analytical benchmark BA solution based on a joint (space/frequency) GP assumption of the shadow fading. The performance depends on the problem and the used features. The results show an LSTM-based solution to outperform other solutions due to its inherent ability to deal with sequential data. Using location and power information in the two bands has consistently shown to be the best BA decision predictor; however, LSTM-based solutions using other information, including delay and AoA, also showed competitive performance.

There are several interesting future research directions. Other labeling solutions can be studied to consider future rate values. Another direction is to expand the features to include raw channel readings and partial observations for fast beam switching (e.g., for low-latency applications). Furthermore, methods to improve the ML solution against practical aspects, such as phone rotation and blockage, e.g., through dataset expansion, are also interesting. One also might utilize other sequential ML-based decision techniques, e.g., reinforcement learning, to solve the BA problem. From the system perspective, multi-BSs dual- and multi-band systems are practically relevant.}

\vspace{-2mm}
\section*{Appendix}
\subsection{GP Based (Additional Details)}
\subsubsection{Exact GP Solution}
To evaluate (\ref{eq:DpIntgSimpl}), since $S^m(t+U)$, $S^c(t+U)$ and the observations in $\mathcal{H}_t$ are jointly normal, it is enough to calculate the conditional mean and variance of $S^m(t+U)$ given $\mathcal{H}_t$ and $S^c(t+U)$, which we denote by $\mu_{m|H^{+}}$ and $\sigma^2_{m|H^{+}}$, respectively. Note that we refer to the set of $\{S^c(t+U) \cup \mathcal{H}_t\}$ by $\mathcal{H}^+_t$. To calculate these quantities we need to define a few vectors and matrices: we use the convention that $\Sigma_X$ denotes the covariance matrix between the elements of a set of random variables $\mathcal{X}$. We also denote $\Sigma_{X,y}$ as the cross covariance vector between $\mathcal{X}$ and a random variable $Y$, and $\Sigma_{y|X}$ is the variance of $Y$ given a realization of elements of $\mathcal{X}$. For instance, $\Sigma_H$ is a $(Q+1)\times (Q+1)$ covariance matrix of the shadowing observations, $\Sigma_{H,b}$ is a $(Q+1)\times 1$ vector that represents the covariance between the shadowing in band $b$ at time $t+U$ and data in set $\mathcal{H}_t$. We use a similar subscript convention for the means, where $\mathbf{m}_X$ refers to the vector of individual mean values of $\mathcal{X}$, and $\mu_{y|X}$ refers to the mean of $Y$ given $\mathcal{X}$. Then it is easy to verify that \cite{leon2017probability}:
\begin{align}
\mu_{m|H^+} = \mu_m + \Sigma_{m,H^+} \Sigma_{H^+}^{-1} (\mathbf{k}-\mathbf{m}_{H^+}) = \Sigma_{m,H^+} \Sigma_{H^+}^{-1} \mathbf{k},
\end{align}
where $\mathbf{k} = [s, { \rm \boldsymbol{vec}}(\mathcal{H}_t)^\top]^\top$, where ${\rm \boldsymbol{vec}}(\mathcal{X})$ converts the set $\mathcal{X}$ to a vector. Also we have 
\begin{align}
\sigma_{m|H^+} = \sigma_m^2 - \Sigma_{m,H^+}\Sigma_{H^+}^{-1}\Sigma_{m,H^+}^\top.
\end{align}
Similarly, for $S^c(t+U)$ given $\mathcal{H}_t$, we have:
\begin{align*}
&\mu_{c|H} = \mu_c + \Sigma_{c,H} \Sigma_{H}^{-1}(\mathbf{h}-\mathbf{m}_{H}) = \Sigma_{c,H} \Sigma_{H}^{-1} \mathbf{h}, \nonumber \\ & ~~{ \rm and }~~ \sigma_{c|H} = \sigma_c^2 -\Sigma_{c,H}\Sigma_{H}^{-1} \Sigma_{c,H}^\top,
\end{align*}
where $\mathbf{h} = {\rm \boldsymbol{vec}}(\mathcal{H}_t)$. Note that as indicated earlier, to calculate these quantities, we have to know the correlation model as well as the path-loss values. Once the above values are calculated, the quantities in eq. (\ref{eq:DpIntgSimpl}) can be evaluated as follows. 
\begin{align*}
\mathbb{P}( S^m(t+U) \geq \mathcal{V}_2(s) & | \mathcal{H}_t, S^c(t+U) = s) = \nonumber \\ & \mathbb{Q}\big(\frac{\mathcal{V}_2(s)-\mu_{m|H^+} }{\sigma_{m|H^+}}\big),
\end{align*}
where $\mathbb{Q}(.)$ is the Q-function \cite{leon2017probability}, and
$f_{S^c| \mathcal{H}_t}(s) = \frac{1}{2\pi \sigma_{c|H}^2} e^{-\frac{1}{2}\big(\frac{s-\mu_{c|H}}{\sigma_{c|H} }\big)^2}.$ 
 
\subsubsection{Approximate GP Solution }
$\tilde{ R}^b$ has a normal distribution with:
$$\tilde{\mu}_b = \omega_b \log(\gamma_b), ~~~~~ {\rm and} ~~~~~ \tilde{\sigma}^2_{b}= \bigg(\frac{\log(10)}{10}\bigg)^2 \omega_b^2 \sigma_b^2. $$ 
For $\tilde{ R }_D$ ($\tilde{ R }_D = \tilde{ R }^m (t+U)-\tilde{ R }^c(t+U)$), has mean and variance, respectively:
$$\mu_D = \tilde{\mu}_m - \tilde{\mu}_c ~~ {\rm and}~~ \sigma_{D}^2 = \tilde{\sigma}_m^2 + \tilde{\sigma}_c^2 - 2\rho_{m,c}\tilde{\sigma}_m\tilde{\sigma}_c.$$
Furthermore, note that $\tilde{ R }_D$ given $\mathcal{H}_t$ is normally distributed, with mean and variance, respectively:
\begin{align*}
& \mu_{D|H} = \mu_D + \Sigma_{D,H} \Sigma_{H}^{-1} \mathbf{h}, \nonumber \\ &{\rm and} ~~\sigma_{D|H} = \sigma^2_D - \Sigma_{D,H} \Sigma_{H}^{-1}  \Sigma_{D,H}^\top.
\end{align*}
Finally, the decision rule becomes
$\mathbb{Q}\big(\frac{-\mu_{D|H}}{\sigma_{D|H}}\big)\geq 0.5$, which can be shown to be equivalent to 
\begin{align}\label{eq:GPApprule}
\mu_{D|H} \geq 0 \implies \tilde{\mu}_c-\tilde{\mu}_m \leq \Sigma_{D,H} \Sigma_{H}^{-1} \mathbf{h}.
\end{align}
\subsection{The Impact of Uncertainty}
For the ray-tracing environment, we study the impact of noisy features (in the form of uncertainty) on the performance. We consider a few feature combinations. We assume that the noise is defined based on upper limits $\alpha e_{{\rm f}_n}$, where $\alpha$ is a positive integer, and $e_{{\rm f}_n}$ is the constant that depends on the feature ${{\rm f}_n}$ (see Sec. \ref{sec:featF}). The choice is based on the grid resolution, where we have $e_{\rm f_1} = 2$ m (distance error), $e_{f_2} = 0.5^o$ (angle error), $e_{\rm f_5} = 6.5$ ns (delay), and $e_{\rm f_3} = 0.25$ dB (power). The values are chosen based on the half distance between grid points. While the LSTM perform relatively well up to twice the grid point separation, we observe what might be attributed to bias-variance trade-off, as complex networks have better capacity but show larger variance, despite the measures that we took to reduce the overfitting, nevertheless, the LSTM based solution is good for small error values. 
\begin{figure}[!ht]
\centering
\vspace{-0 mm}
 \includegraphics[width=1\columnwidth, trim={0cm 0cm 0cm 0cm},clip]{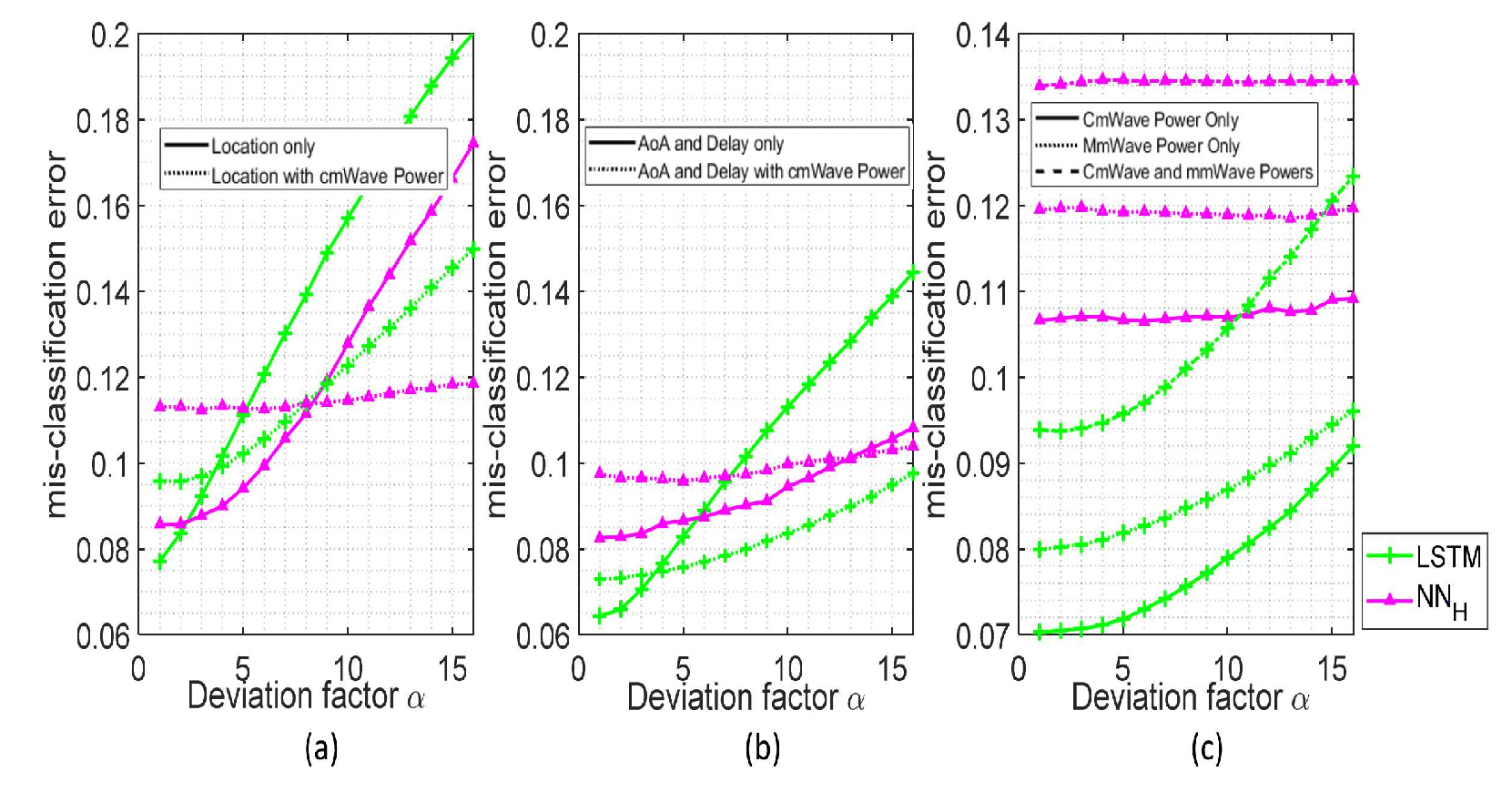} 
  \vspace{-7mm}
\vspace{-0mm}
\caption{The impact of noise on the performance (${\bar{\mathcal{E}}_S}$) when using different features: (a) location (distance and phase), (b) AoA and delay, and (c) omni. powers. For (a) and (b) the features combinations are shown with and without cmWave power.}
\label{Fig:PeVsURT}
 \vspace{-2mm}
\end{figure}

{ In another study (not shown here for space limitation), we trained and tested the LSTM and GR algorithm on one dataset generated with one value of $\nu$ and tested on another dataset generated by another $\nu$ value, $\nu\in\{1,1.9\}$. We noticed a significant impact (the error values are doubled) compared to train and test on the exact $\nu$ value, with more losses when training on $\nu=1.9$. This is somewhat not surprising due to the environment mismatch, and $\nu=1.9$ enforces longer memory that might not be needed for $\nu=1$.}

\subsection{A Worked-out VA Labeling Example}

\begin{figure*}[!htb]
\vspace{-1mm}
 \scriptsize

\centering
\vspace{2mm}

\includegraphics[width=1.6\columnwidth, trim={0cm 0cm 0cm 0cm},clip]{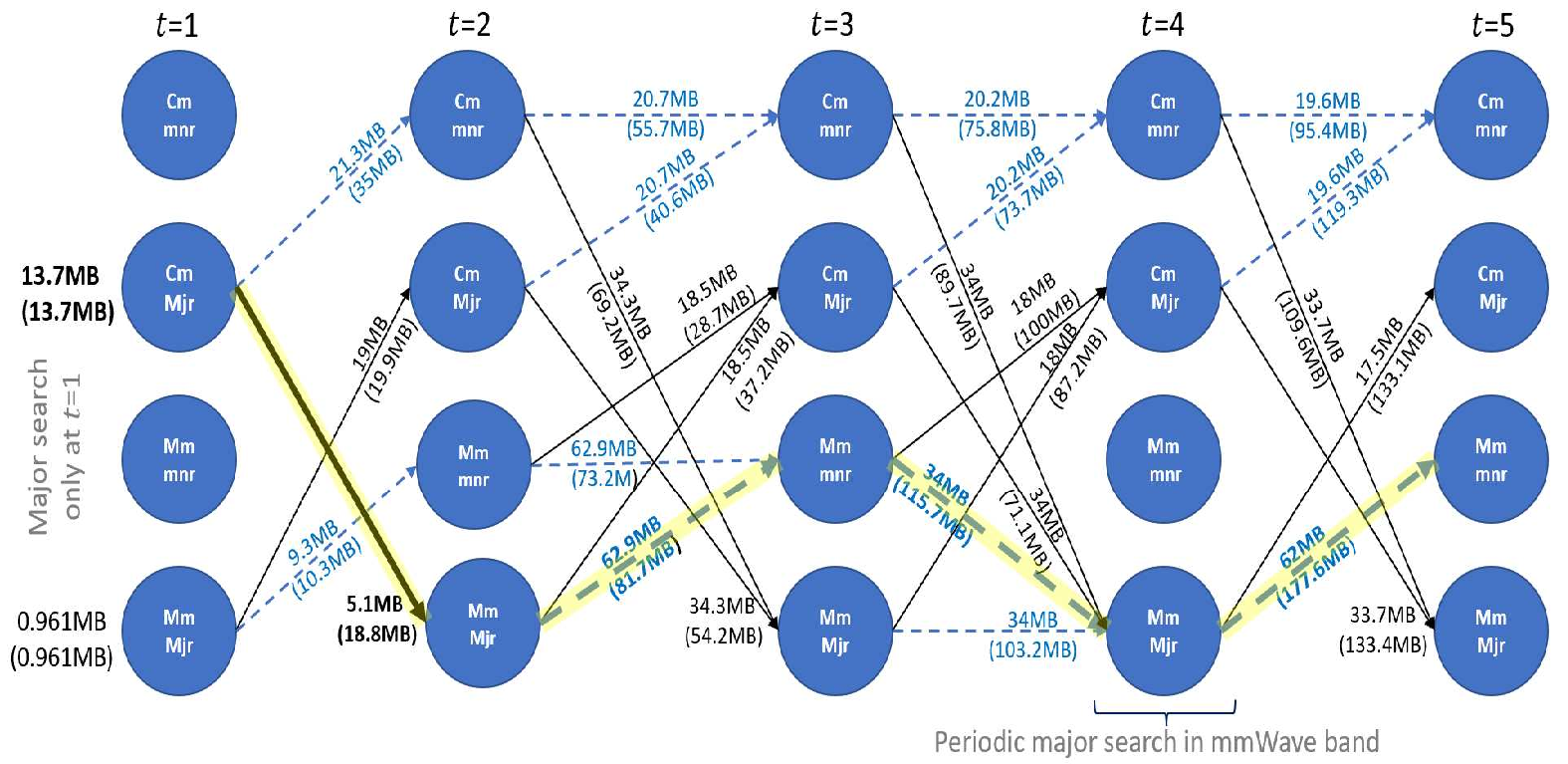}
\vspace{-0mm}
\caption{ An example of labeling a sequence of length five time-frames. Two states in each band. The tellies diagram format is similar to Fig. \ref{fig:VAstates}. On each connecting line, we show the transferred data (after discounting the time needed for beam-search) in that state, followed by the accumulated data up to that state in parentheses. The selected route is shown with thick lines (also highlighted), and the data is marked with bold font.}
\label{Fig:VAExample}
 \vspace{-5mm}

\end{figure*}

In Fig. \ref{Fig:VAExample}, we show a worked-out labeling example to the first five time steps of one of the sequences in the ray-tracing dataset. In the chosen sequence, no outage or change in the best beam combination occurs throughout the five time-steps (chosen for clarity of the presentation), for more details please refer to Fig. \ref{fig:mnrMajr}. The trellis diagram shows the possible transferred data in one time-step and the total received data up to that state. The solution assumes major searches at the beginning. In this example, $T^b_s$ in (\ref{eq:ratediscount}) (i.e., $T^b_M$ and $T^b_m$) and  (\ref{eq:TbM}) are calculated based on the previously given parameters in Sec. \ref{subsec:ImpMIMOdetails}. Note here we use the actual data transfer (i.e., multiply (\ref{eq:ratediscount}) by $T_f = 2.6$ s). Due to space limitations, we do not give the actual data rates per band as they can be derived based on the above and the values shown in Fig. \ref{Fig:VAExample}. 

The chosen sequence in Fig.\ref{Fig:VAExample} is $\big\{${\em cmWave} (Cm Mjr), {\em mmWave} (Mm Mjr), {\em mmWave} (Mm mnr), {\em mmWave} (Mm Mjr), {\em mmWave} (Mm mnr)$\big\}$, which achieves a total transferred data of $177.6$ MB. Based on the operating bands of this sequence, the corresponding labels are $\{0, 1, 1, 1, 1\}$. A greedy instantaneous labeling results in $\{0, 0, 1, 1, 1\}$ and a total transferred data of $165.2$ MB.

\bibliography{reportBib}

\end{document}

%% file: nIntroduction.tex
The large available bandwidths in the millimeter-wave (mmWave) frequency band can support the high data rates required for many emerging applications in next generation wireless networks (5G and beyond). However, the hostile propagation conditions at high frequencies restrict its utilization. Compared to the centimeter-wave (cmWave) band,\footnote{In a slight abuse of notation, we call here the sub-6GHz band the cmWave band, and 24-100 GHz the mmWave band. This is inspired by the current 3GPP and WiFi frequency ranges.} signals in the mmWave band suffer from higher attenuation, higher diffraction loss, and are more susceptible to blockage, which reduces the reliability of the communication systems \cite{shafi2018microwave}, a required criterion for seamless user experience. The {\em joint} utilization of the two bands enhances the coverage, system reliability, and achievable data rates. This has been confirmed with numerous research studies, where the outage probability and throughput are shown to improve significantly \cite{burghal2018Rate,8790809,cheng2020joint, 7493676,8198818}. Thus, due to these characteristics of the two bands, \emph{both} are indispensable components for future wireless networks \cite{andrews2014will,shafi20175g,dahlman20205g,dang2020should}.

Communications exploiting dual bands can be realized in different ways \cite{chandra2015cogcell} \cite{lien20175g}. For instance, the cmWave band can be used for the control plane while the mmWave band is used for the data plane. Alternatively, both bands can be used for both planes. In fact, these scenarios were also among the various possible architectures that were proposed for the initial deployments of the non-stand alone (NSA) mode of the 5G mobile networks, i.e., through coexistence of the Long Term Evolution (LTE) and the 5G NR\cite{dahlman20205g}. However, the \emph{simultaneous} usage of the two bands might not be practical due to a number of limitations at the network side and/or at the UE side, such as limited processing capabilities, constraints on transmission power, etc. Thus, depending on the underlying BA scenario, the BS\footnote{We use the term BS as a generic expression for the gNodeB in 3GPP NR, or an Access Point with both 802.11ac/ax and 802.11ad/ay in WiFi. Furthermore, the dual-band capabilities of the "BS" here could be realized with co-located BSs.} has to assign the UE to one of the two bands based on the observed channels, such as in the initial channel access scenario, or it has to sequentially switch the communication between the two bands as the UE moves, i.e., switching to the mmWave band whenever it is available or to the cmWave band when the mmWave band suffers a blockage or other bad propagation conditions. We refer to the first problem as a \emph{one-shot} BA, and the second as a \emph{sequential} BA. We here focus on the latter problem as it can be viewed as a generalization of the one-shot, which was addressed preliminary in our previous work \cite{burghal2018band}.

In general, the BA problem is challenging because simultaneous observations of the two bands are not usually available to the BS. One solution can be through using a frequent "measurement gap" to send training signals (e.g., Channel Status Information Reference Signal (CSI-RS) in NR) over the two bands; however, the resulting overhead and/or delays reduce the overall throughput of the system. The problem is exacerbated with the needed signaling to schedule the switching and perform the handover between frequency bands, or when the UE and/or BS use multi-antennas, where beam search is needed to align the beams. As an alternative solution, the system could rely on the correlation and the joint characteristics of the two bands \cite{gonzalez2017millimeter,aliestimating}. For the sequential BA, where the goal is to choose the band for a future time frame, the joint characteristics of the bands are intricate, as the channel realizations and the relation between the two bands change and decorrelate over time. Alternatively, the BS can utilize partial information, such as the channel state in one band or the UE's location, together with some "prior knowledge" of the cell environment to solve the BA problem; this is a promising approach as the side information could reveal the underlying structure of the environment.\footnote{Note that the same concepts have been used in different wireless applications, such as localization, where the mapping between the channel information and location is used in fingerprinting approaches \cite{burghal2020comprehensive}.} However, realizing this solution is challenging as it needs to rely on the joint characterizations of the temporal evolution of the two bands and the proper utilization of such (possibly) non-homogeneous information. In this paper, we propose solutions based on ML. For the extensive study of the joint characterization of the channels, the readers are referred to \cite{sangodoyin2018Joint, burghal2019dual}, and the references therein.

ML provides powerful techniques that can capture complex relations between the input data (features) and the output values (labels). Motivated by the remarkable success of ML in various fields, the wireless communication community has shown an increased interest in ML-based solutions for channel coding, estimation, channel modeling, to name a few, see, e.g., \cite{chen2019artificial,jiang2016machine,zhang2019deep} and references therein. The reported results are promising; ML-based solutions are able to provide competitive performance for problems where optimal solutions are known, e.g., using multi-layer Neural Networks (NNs) for decoding in AWGN channel \cite{o2017deep}, indicating that ML may also be applied to problems where traditional methods have failed or where the environment is too complex. For instance, Ref. \cite{farsad2018neural} demonstrated the efficacy of ML-based detection over a molecular system where the channel characteristics are difficult to model. 

Similarly, for the simplified BA problems that can be handled analytically (one-shot BA in a Gaussian stochastic channel), ML solutions show comparative performance to the optimum closed-form solutions \cite{burghal2018band}. Motivated by this, and due to the complexity of the sequential BA, we propose to solve the BA problem using supervised ML techniques with \emph{various} features combinations. The solutions are based on feedforward NN and recurrent NN (the Long Short Term Memory (LSTM)). We consider systems with omnidirectional and directional antennas to capture different aspects of the BA problem. We use the omnidirectional system to study the basic problem and derive analytical benchmarks in a synthesized stochastic environment. In particular, with the common assumption that the shadow fading is normally distributed (in decibel (dB) scale), the analytical solution maps the observed GP to the BA decision (based on the probabilities of the achieved rates). The directional systems can be viewed as a generalization, where the system needs to identify the best direction to use at the link ends, and incorporate the increased overhead from band switching. As we discuss below, acquiring labeled data when using directional antennas is challenging. To solve this, we adopt basic principles of the 5G-NR beam searching architecture, and use the VA to propose an efficient (heuristic) labeling algorithm that takes the accumulated rate into account. In addition to the synthetic dataset that we use to compare the ML-based to the analytical benchmark (GP) solutions (in omnidirectional case), we utilize ray-tracing to generate a dataset for omni- and directional cases. Finally, note that in practice, directional communication is usually achieved via analog beam-forming of multi-antennas (essential for mmWave communication), thus in this paper, we also refer to the directional case as Multi-Input Multi-Output (MIMO) case, while for the omnidirectional one as Single-Input Single-Output (SISO) case.

%% file: nPriorWork.tex
There have been recent studies that considered the {\em interplay} between cmWave and mmWave bands. Refs. \cite{Nitsche2015Steering,hashemi2017out,aliestimating} utilize the angular correlation in the two bands to provide a coarse estimate of the Angle of Arrival (AoA) at mmWaves based on the AoAs in the cmWave band, which can be used to reduce the beam-forming complexity at the mmWave band. Ref. \cite{ali2018spatial} studied the covariance matrix translation between the two bands. For joint communication in the two bands, \cite{hashemi2017out} proposes a two-queue model to assign data to each band such that delay is minimized and throughput is maximized. Ref. \cite{semiari2017joint} considers the downlink resource allocation in a network with a small cell BS, where the BS aims to assign the UE or services to the resources in the two bands. 

The BA can be viewed as a handover process between two co-located BSs with different frequencies. Refs. \cite{chergui2018classification,mismar2018partially,alkhateeb2018machine} used ML approaches to address the handover and switching between BSs that may use different frequency bands. In \cite{mismar2018partially}, the authors use ML to improve the success rate in the handover between two co-located cells in different bands, their implemented ML classifier uses the prior channel measurements and handover decisions within a temporal window to predict the success of the handover. Ref. \cite{chergui2018classification} introduces an uplink (ULink)/downlink (DLink) decoupling concept where the central BS gathers measurements of the Rician K-factor and the DLink reference signal received power for both bands, and trains a non-linear ML algorithm that is then applied to the cmWave band data to predict the target frequencies and BS that can be used for the ULink and DLink. Ref. \cite{alkhateeb2018machine} uses a gated recurrent NN to predict handover status at the next time slot given the beam-sequence, where the BS uses the sequence of previously used {\em beam-forming} vectors as input to the ML scheme. Different from these works, we use various sets of features and several ML algorithms for omnidirectional and directional systems in two different environments, this help identifying the limitations of the solutions and the impact of the features. We also consider an analytical solution based on GP for the BA problem, which allows us to benchmark the ML solution. Furthermore, we propose an efficient labeling scheme, i.e., providing a key enabler for the supervised ML-based BA solutions. 

Channel state prediction using GP or ML was considered in several works such as \cite{riihijarvi2018machine,wang2007predicting,wang2017spatiotemporal,muppirisetty2016spatial}, where Refs. \cite{riihijarvi2018machine,muppirisetty2016spatial} use GP to predict the shadowing values in the network based on collected drive tests, while Refs. \cite{riihijarvi2018machine,wang2007predicting,wang2017spatiotemporal} use regression ML techniques to predict the channel state. Using ML to predict unobserved channel features was also considered in \cite{chen2017remote,Navabi2018Predicting}; in \cite{Navabi2018Predicting}, the authors use NNs to predict the AoA, and in \cite{chen2017remote}, the authors utilize the observed channel state information in a central BS to predict the optimal beam direction in local BSs. In Ref. \cite{burghal2019machine}, the authors use beamformed CSI as input to a recurrent NN to track the AoA. However, these works focus on a single band and use mainly a regression framework, while in this paper, we solve the BA in two bands as a classification problem using several related feature combinations.

 In our prior work\cite{burghal2018Rate}, we utilize the GP assumption to propose and analyze {\em linear prediction based} BA solution in an omnidirectional communication system\cite{burghal2018Rate}. However, in this work, the objective of the GP benchmark is to minimize the misclassification probability. In \cite{burghal2018band}, we discussed the one-shot BA.

%% file: nContribution.tex
The contributions of this work span different aspects of the BA problem, which can be summarized as follows.
\begin{itemize}
    \item Formulate the BA in dual-band omnidirectional and directional systems as binary classification problems and propose supervised ML solutions based on recurrent NN (LSTM) and different benchmarks. 
    
    \item To enable supervised ML solutions, we study the labeling techniques for the sequential BA. The labels are chosen such that the sum rate is maximized. \begin{itemize}
        \item For a {\em simplified} SISO {\em (omnidirectional communication)}, where the switching cost is ignored, the labels can be based on the pairwise rates.
        \item For MIMO systems, where the multi-antenna are used for beam-forming at both sides, we incorporate some basic assumptions from 5G-NR to formulate the structure of the beam search and its periodicity. Then we construct a trellis diagram to capture the switching cost and propose a labeling scheme that captures the sequential BAs.
    \end{itemize}
    
    \item In the omnidirectional systems, we utilize the normal distribution of the shadow fading (in a stochastic environment) to propose an analytical benchmark solution (dubbed as GP-based solution), and derive both exact and approximate versions of the solution.
    
    \item For a fair assessment, we study the performance against simpler ML solutions, and study the performance under several features combinations in stochastic and ray-tracing environments. 

\end{itemize}

 The remainder of the paper is organized as follows. Sec. \ref{sec:overview} introduces the basic system model, formulates the sequential BA problem, and introduces the proposed solutions. Sec. \ref{sec:DataAspects} highlights the two environments and the method used to construct the sequences. A detailed discussion in the section is dedicated to the proposed labeling method. Sec. \ref{sec:MLsolution} elaborates on the ML solution. Sec. \ref{sec:GPBasedSol} summarizes the GP-based benchmark solution for omnidirectional SISO systems. The performance of SISO systems in a stochastic environment is studied in Sec. \ref{Sec:StochEnvDL}. The performance of the solutions for the omnidirectional SISO and the directional MIMO systems in a ray-tracing environment are studied in Sec. \ref{Sec:RTEnvDL}. Finally, Sec. \ref{Sec:ConcDL} provides concluding remarks and suggested future directions.

%% file: BasicSystemModel.tex
We consider a dual band cellular system, where the BS and the UE can operate in two frequency bands with center frequency $f_b$ and bandwidth $\omega_b$ in band $b \in \{c,m\}$, where $c$ and $m$ refer to the cmWave and the mmWave bands, respectively. When using multi-antennas, without loss of generality, in band $b$ the BS and the UE use beamforming codebooks $\mathcal{K}^b$ (of size $K^b$) and $\mathcal{M}^b$ (of size $M^b$), respectively. Due to a number of practical limitations of the UE, we assume that data transmission occurs in a single frequency band at a time. The BS controls the band selection process, using some observations about the channel and prior knowledge to choose the band that results in the highest data rate, more details about the band selection metric are provided later in this section. To focus on the basic problem, we consider a single user case, i.e., no scheduling or interference is considered; the multi-user case is left for future work. 
 
It is well established that the small scale fading in the two bands are independent due to the large frequency separation; furthermore, modern diversity techniques mostly eliminate its impact \cite{ProfMolischText}. In contrast, large scale parameters vary relatively slowly over time and maintain time and frequency correlation, making it possible to utilize information over frequency and time (space) and thus make switching decisions. {Note also that large-scale parameters are reciprocal in ULink and DLink for both time-domain and frequency-domain duplexing systems as long as the duplexing distance is smaller than the stationarity time or bandwidth, respectively; this condition is fulfilled for almost all practical systems. For this reason, the subsequent discussion is valid for both link directions, and we assume that the BS can acquire the channel state information about the large-scale parameters without additional overhead.} 
 
Similar to \cite{burghal2018Rate} we define a time-frame as a sequence of $T$ time slots (data units), each time-frame is indexed with $t$, see Fig. \ref{Fig:ProbDesc}. On a dB scale, we can write the Signal to Noise Ratio (SNR) of the received signal in band $b$, during time-frame $t$ \cite{ProfMolischText}:
\begin{align}\label{eq:DPSNR}
{\rm SNR}^{b}_{m,k}(t) = P^{b}_{m,k} +\zeta^b_{m,k}(t) - N^b_{m},
\end{align}
where we assumed the UE and the BS use the $m^{\rm th}$ and the $k^{th}$ beamforming codeword from $\mathcal{M}^b$ and $\mathcal{K}^b$, respectively. $P^{b}_{m,k}$ is the Effective Isotropically Radiated Power (EIRP), $N^b$ is the noise level, and $\zeta^b(t)$ captures the large scale variation in band $b$ that varies as the UE moves. Then using capacity-achieving transmission, we can write the rate in band $b$ as
\begin{align}\label{eq:DPrate}
R^b_{m,k}(t) = \omega_b \log\big(1+10^{0.1 {\rm SNR}^{b}_{m,k}(t)}\big).
\end{align} 
The BA procedure and the detailed description of the observations and prior knowledge depend on the scheme and the setup. In general, the BS used the observations to produce the soft decision $\tilde{\mathcal{D}} \in [0,1]$, which it then uses to make the BA decision $\mathcal{D}\in \{0,1\}$, where "0" and "1" refer to the data transmission in the cmWave and mmWave band, respectively.

%% file: ProblemDesc.tex
\begin{figure}[t]
\centering
\vspace{2mm}
\includegraphics[width=1\columnwidth, trim={0cm 0cm 0cm 0cm},clip]{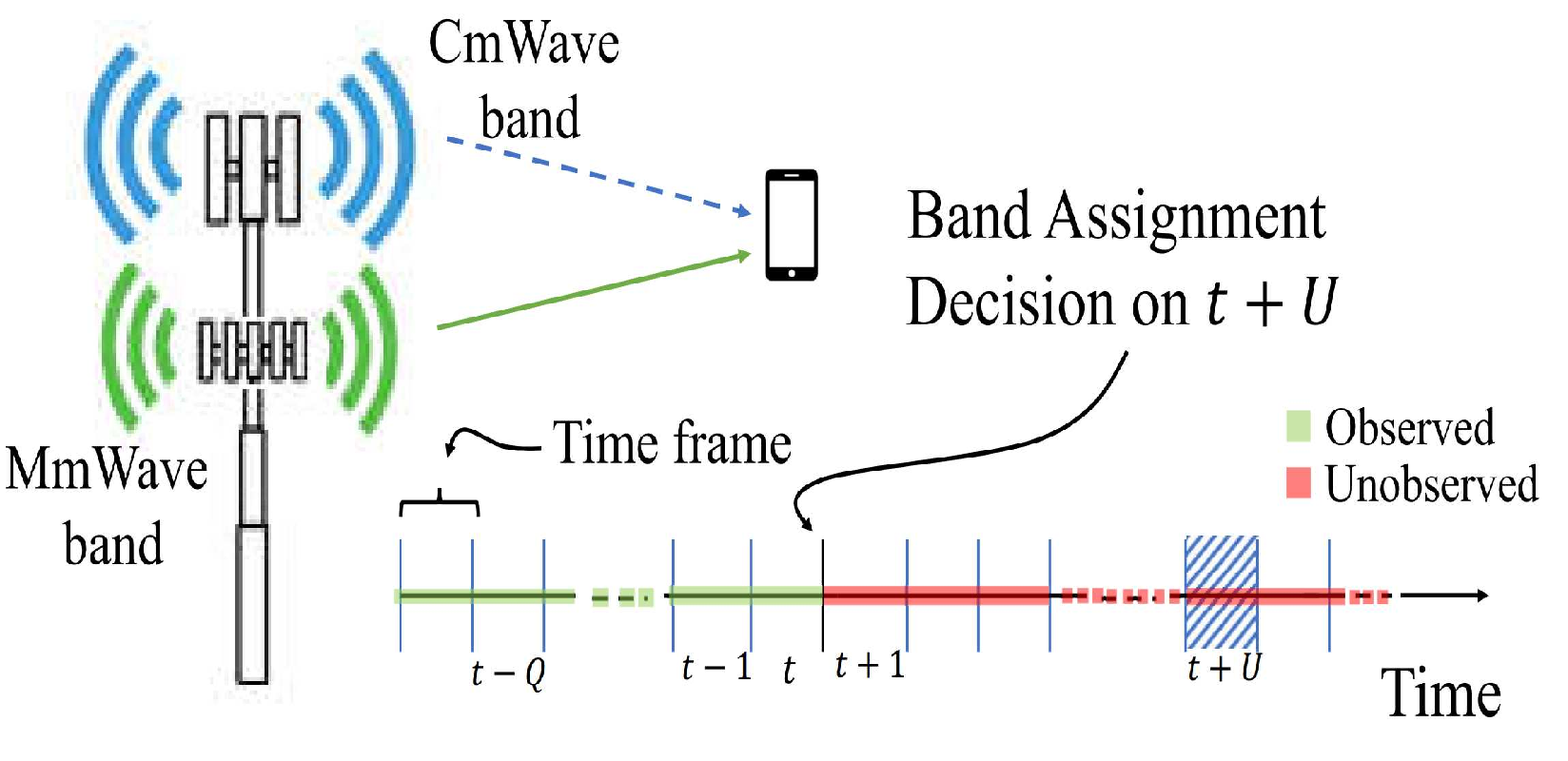}
\vspace{-2mm}
\caption{Description of the problem and the time series of observations and decision.}
\label{Fig:ProbDesc}
 \vspace{-6mm}
\end{figure}

The BA problem can be summarized as follows, see Fig. \ref{Fig:ProbDesc}. As the UE moves, the BS uses the current and the \emph{previous} observations to predict the best band \emph{after} $U$ time frames, i.e., the BS makes sequential decisions. The used observations depend on the approach, but may include the observed power (or rate) in one of the two bands, the UE location, the delay and the AoA of the dominant Multi Path Component (MPC) at frame $t$. There are many applications to the sequential BA as part of the resource allocation, e.g., when the network has to plan what resources to use in advance.
 
Depending on the type of communication, we consider two setups.
\begin{itemize}
    \item Omnidirectional SISO system, ({\em SISO} for short), or equivalently, $ {K}^b  =  {M}^b  = 1$.
    \item Directional communication with analog beamforming in MIMO system, ({\em MIMO} for short). In such system the BS and the UE have to search for the best beam combination, which increases the overhead and reduces the spectral efficiency.
\end{itemize}
 We make this distinction since in SISO: (i) The GP modeling assumption of shadow fading is reasonable, (ii) the derivation of an analytical benchmark solution is tractable, and (iii) it is clearer to describe the different aspects and factors that influence the BA.

%% file: PerformanceMetric.tex
The ultimate goal of the proposed BA is to maximize the data rate.
In the sequential BA, the sum rate is the preferred metric, so the impact of the sequential switching cost has to be taken into account. This cost could be due to the time needed for beam search and alignment in the new band. One way to capture that is through the normalized total rate loss $\bar{\mathcal{R}}_{X}$. Let ${R}^*_j(t)$ be the rate at time instant $t$ when the optimal labeling is used for sequence $j$ (the discussion of how to label the data is deferred to Sec. \ref{sec:labeling}). For {\em a given scheme}, we can define the achievable rate as $\Tilde{R}_{j}(t) : \{R^b_j(t): b = m ~{\rm if}~ \mathcal{D}_{{\rm s}j}(t) =1,~{\rm or}~  b = c ~{\rm if}~ \mathcal{D}_{{\rm s}j}(t) =0\}$, where $\mathcal{D}_{{\rm s}j}(t)$ is the decision by the scheme for sequence $j$ at $t$. The rates $R^b_j(t)$ are calculated based on eq. (\ref{eq:DPrate}) while taking into account the usable time period (after switching) along with the possible restrictions on the set of the beamforming vectors $m$ and $k$. Then the normalized rate loss is given by
 
\begin{align} \label{eq:sumRate}
    \bar{\mathcal{R}}_{X}  = \frac{1}{N_{{\rm s}X}} \sum_j^{N_{{\rm s}X}} \bigg| \frac{\sum_{t}^{T_j} \Tilde{R}_j(t) - \sum_{t}^{T_j}  {R}^*_j(t)}{\sum_{t}^{T_j} R^*_j(t)}\bigg|, 
\end{align} 
where $N_{{\rm s}X}$ is the number of {\em sequences} in data set $X$, $T_j$ is the length of the $j^{\rm th}$ sequence.

Since the BA is formulated as a classification problem, we use the mis-classification errors as a performance metric. We here adopt the average number of BA errors $\bar{\mathcal{E}}_X$.\footnote{Since some of the data points are correlated we rely on an argument similar to the one we gave in \cite{burghal2018Rate}.} Defining $N_{X} = \sum_j^{N_{{\rm s}X}} T_j$ as the total number of time instances, then 
\begin{align}\label{eq:error}
\bar{\mathcal{E}}_X =\frac{1}{N_{X}} \sum_j^{N_{{\rm s}X}}\sum_t^{T_j} |\mathcal{D}_{{\rm s}j}(t)-\mathcal{L}_{{\rm s}j}(t)| =\frac{1}{N_X}\sum_i^{N_X} |\mathcal{D}_i-\mathcal{L}_i|,
\end{align} 
where $\mathcal{L}_{{\rm s}j}(t)$ is the label at time frame $t$ of the $j^{\rm th}$ {\em sequence} (more details are given in the labeling subsection \ref{sec:labeling}). Since the data points over which we evaluate the performance may belong to different sequences, we may drop the sequence dependency and thus omit the subscript ${\rm s}$, the index $t$, and replace the sequence index $j$ with \emph{instant} $i$. Below we may drop the subindices the context is clear. The $\bar{\mathcal{E}}_X$ can be interpreted as probability of BA error. Triggered by (\ref{eq:sumRate}), one might choose to calculate the average number of mis-classifications {\em per sequence}, however, we adopted (\ref{eq:error}) as the two are approximately equivalent for a large number of sequences and it allows for simplifications in our analysis. 

%% file: SolutionOverview.tex

\subsubsection{Learning Based Solutions} \label{subsec:DPIntroLrn}
In any given frame $t$ the BS has to take one of two decisions: to use cmWave band ($\mathcal{D}_{{\rm s}j}(t)=0$) or to use the mmWave band ($\mathcal{D}_{{\rm s}j}(t)=1$), which can be viewed as a binary classification problem. Several ML models can be used to solve such problems. While our focus will be here on NN-based solutions, we also consider a simpler ML model to provide fair comparisons of the proposed solutions. 

In particular, we use a deep network with FC-NN and LSTM layers. We also use a multi-layer NN and GR with historical data for comparison, the history being a window of the last $Q$ observations (see Fig. \ref{Fig:ProbDesc}). The observed features depend on the setup, and may include the location information, and/or the observed omni-directional power from one or both frequency bands (needed for the analytical solution).\footnote{The omni-directional power in both bands is just one of many possible features, in practice this can be realized using passive power sensing (RSSI sensors) \cite{ProfMolischText}.} 

For the ML-based solutions, we denote the set of observed features in time frame $t$ by $ \mathcal{F}_t$. Details of the chosen approaches, training, etc. are presented in Sec. \ref{sec:MLsolution}. Note that we apply the learning solutions to both SISO and MIMO settings.
\subsubsection{Gaussian Process Based Solutions}\label{subsec:GPsolModl}
In this approach, we adopt an analytical solution and apply it, for tractability, to the SISO system only. Here we assume that the $\zeta^b(t)$ in (\ref{eq:DPSNR}) consists of path-loss $P_{\rm L}^{b}(t)$ and large scale fading (shadowing) $S^{b}(t)$, i.e.,
\begin{align} \label{eq:DPZeta}
    \zeta^b(t) = -P_{\rm L}^{b}(t)+S^{b}(t)
\end{align}
We assume that the BS knows the channel model and statistics, and it can estimate $P_{\rm L}^b(t)$, either using empirical models or prior knowledge of the environment. Also we assume that the shadowing is a stationary GP in space (time) and frequency with mean $\mu_b = 0$ and standard deviation $\sigma_b$ in band $b$, i.e., $S^b(t) \sim \mathcal{N}(0,\sigma_b^2)$, and $S^b(t)$ and $S^{b'}(t')$ are jointly normal with correlation function ${ \bf Cov}(S^{b}(t),S^{b'}(t'))$. An example of the correlation model and further discussion is provided in Sec. \ref{sec:GPBasedSol} and in \cite{burghal2019dual}. Note that assuming $S^b(t)$ is Gaussian on a dB scale matches many measurement campaigns \cite{ProfMolischText}, but it may not always hold in practice. Still, we rely on it along with the joint Gaussian assumption over frequency for simplicity and mathematical tractability. To emphasize the fact that the rate is a random quantity, we use (\ref{eq:DPZeta}) to rewrite (\ref{eq:DPSNR}):
\begin{align}\label{eq:rate}
R^b(t) = \omega_b \log(1+ 10^{ (P^{b}_{\rm tx} - P_{\rm L}^{b}- N^b_0)  }  10^{ 0.1 S^b(t)}).
\end{align} 
Note that the subscripts for beamforming codewords are not used as this approach is used only for SISO systems.  
To simplify the notation we sometimes omit the time index for $S^b$ and $R^b$ when the meaning is clear from the context. 

In time frame $t$ the GP-based solution uses the SNR observations in both bands (along with the path loss estimates) of the current and the last $Q$ time frames to extract $S^b$ values, then it predicts the BA decision in time frame $t+U$ based on the {\em probability} that $R^b(t+U) > R^{b'}(t+U)$.

\subsection{Remarks}
{\em Remark 1}: In both solutions, when given a soft decision $\tilde{\mathcal{D}}$ (the output of the ML or probability in GP-based solution), the BS can use a threshold $\gamma_{\rm T}\in [0,1]$ to map $\tilde{\mathcal{D}}$ to ${\mathcal{D}}$. We assume that ${\mathcal{D}} = 1$ when $\tilde{\mathcal{D}}>\gamma_{\rm T}$. We can, in general, choose the $\gamma_{\rm T}$ that results in the best performance, however, we employ $\gamma_{\rm T} = 0.5$, more details are provided in the simulation sections. 

{\em Remark 2}: In a simplified setting, for the one-shot problem, the BS observes ${\rm SNR}^b(t)$ to extract $S^b(t)$ and then makes a BA decision based on the probability that $R^b(t) >R^{b'}(t)$.

{\em Remark 3}: For the GP-based solutions we refer to the set of observations at time $t$ as set $\mathcal{H}_t$. In general, it is easy to observe that this approach uses power (rate) and distance information; it also uses previous measurements to acquire the statistics of the environment. However, it is difficult to directly incorporate other features. Thus we consider two different environments, one of which matches the GP channel model.

{\em Remark 4:} Note the GP-based solution in this paper is based on the classical wireless communication procedure, estimating path-loss, covariance parameters, etc. Although, the mathematical structure might resemble an ML algorithm, GP-Classifier \cite{murphy2012machine}, the input features and the procedure to estimate (train) the parameters are different, making them two distinct approaches.

%% file: Stoch_TheEnv.tex
In order to generate the channel realizations in the two bands, we use a correlation model proposed in \cite{burghal2019dual}. The covariance between shadowing values at time instants $t$ and $t'$ and in frequency bands $b$ and $b'$ is 
\begin{align}\label{eq:xcorr}
{\bf  Cov}\big(S^b(t),& S^{b'}(t')\big)  = \\ \nonumber  & \rho_{b,b'} \sqrt{{\bf  C}\big(S^b(t), S^b(t')\big) {\bf C}\big(S^{b'}(t), S^{b'}(t')\big)},
\end{align} \vspace{0 mm}
where $\rho_{b,b'}$ is the inter-band correlation coefficient, and
\begin{align}\label{eq:Dpcov}
{\bf  C}\big(S^b(t), S^b(t')\big) = \exp^{-\big(\frac{\Delta(t,t')}{d^b_{\rm dcor}}\big)^\nu}\sigma_b^2,
\end{align}
where $\Delta(t,t')$ is the displacement (in meters) between the location of the UE at times $t$ and $t'$, $d^b_{\rm dcor}$ is the shadowing decorrelation distance in band $b$ (in meters), the real coefficient $\nu >0$ is a decay exponent \cite{szyszkowicz2010feasibility}, values for $\nu$ in $(0,2]$ have been previously used \cite{szyszkowicz2010feasibility}. Note that with $\nu = 1$ (\ref{eq:Dpcov}) is equivalent to the popular Gudmundson correlation model\cite{burghal2019dual}; with this value, we observed that the schemes show small dependency on prior observations, which may not reflect practical environments, thus we consider two values of $\nu$ in the sequential problem. We assume that the path-loss follows a breakpoint path-loss model\cite{ProfMolischText}, with a break distance $d_{\rm break}$ and a propagation exponent $2$ for $d \leq d_{\rm break}$ and $\epsilon$ for $d > d_{\rm break}$. Table \ref{Tab:SimEnvrionStc} summarizes the values used for generating the data sets. 
\input{Tables/EnvParameters.tex}

%% file: Tables/EnvParameters.tex
\begin{table*}[!htb]
    \begin{minipage}{0.5\linewidth}
     \vspace{1.5mm}
    \scriptsize
\centering
 \begin{tabular}{|c|c|}
 \hline
\bf Variable & \bf Band c/m  \\
 \hline
\bf $f_b$  & 2.5/28 GHz \\
 \hline
 Bandwidth $\omega_b$ & 10/100 MHz \\
 \hline
\bf $P^{b}_{\rm tx}$ & 15/22 dBm \\
 \hline
\bf $\epsilon$ & 4 \\
 \hline
 \bf $d_{\rm break}$ & 50 m \\
 \hline
\bf $d_{\rm dcor}$ & 25/24\\
 \hline
\bf $\sigma_b$ & 5/7 dB\\
 \hline
\bf  $\rho_{m,c}$ & 0.75\\
 \hline
\bf  $\nu$ & \{1,1.9\}\\
 \hline
   Noise Spectral Density & -174 dBm/Hz\\
 \hline
 \end{tabular}
 \vspace{-1mm}
  \caption{Stochastic channel simulation configurations}
 \label{Tab:SimEnvrionStc}
    \vspace{-7mm}
    \end{minipage}\hfill
       \begin{minipage}{0.45\linewidth}
 \vspace{0mm}
 \scriptsize
\centering
 \begin{tabular}{|c|c|}
 \hline
 \bf Variable &\bf Band c/m \\
 \hline
$ f_b $ & 2.5/28 GHz \\
 \hline
  Ant. Pattern & Isotropic \\
 \hline
 Ant. Polarization & Vertical \\
 \hline
 $ {P^{b}_{\rm tx}}$ & 15/30 dBm \\
 \hline
  BS height & 45 m \\
 \hline
  MS height & 2 m \\
 \hline
  Max. Diffraction & 2/1\\
 \hline
  Max. Reflection & 10\\
 \hline
 \end{tabular}
 \vspace{2mm}
  \caption{Ray-tracing simulation configurations.}
   \vspace{-6mm}
 \label{Tab:SimEnvrionPara}
    \end{minipage}\hfill
   \hfill
\end{table*}

%% file: RT_TheEnv.tex
\begin{figure}
\centering
\vspace{3mm}

 \includegraphics[width=0.655\columnwidth, trim={-2cm 2cm 15cm 2cm},clip]{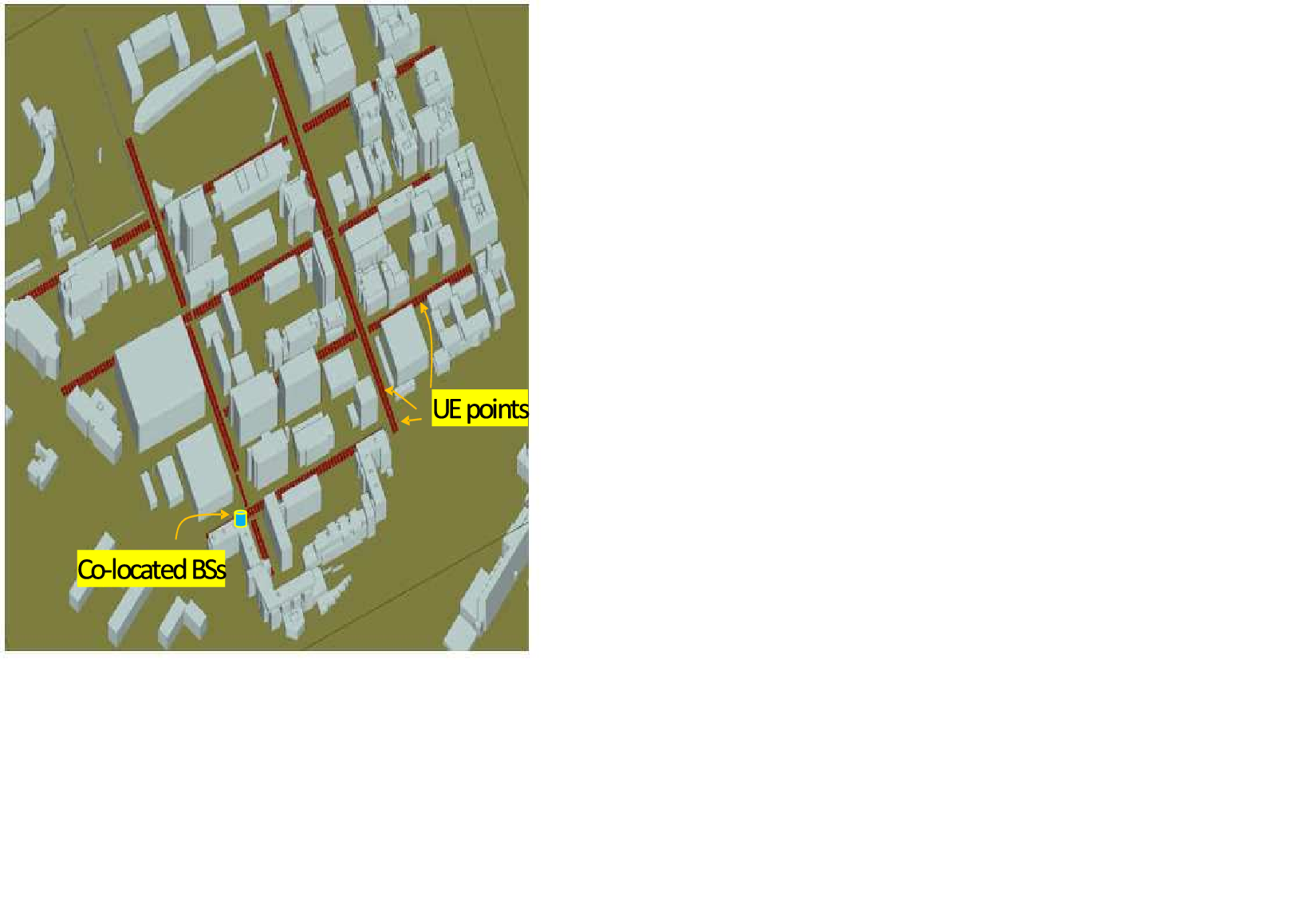}
\vspace{-10mm}
\caption{ Ray-tracing simulation environment. The co-located BSs are above the rooftop. Gray objects represent the buildings. The green 3D polygons denote foliage with different densities.}
\label{Fig:Ray-tracing_USC}
 \vspace{-4mm}
\end{figure}

To assess the performance in a more realistic setting, we simulate the propagation channel in a campus environment by means of a commercial ray-tracing tool, Wireless InSite \cite{wiweb}. The input to the ray-tracer includes the 3D models of the buildings, the characteristics of the building materials and models of foliage. 
The output is a list of parameter vectors that contains the power, propagation delay, the Angle of Departure (AoD) and the AoA, for each MPC. Simulation results have been compared to measurements in a variety of settings and good agreements were observed\cite{wiweb}. This simulation has been conducted based on the model of the University Park Campus, University of Southern California, which is shown in Fig. \ref{Fig:Ray-tracing_USC}-(a). The detailed simulation configurations are listed in Table \ref{Tab:SimEnvrionPara}. The simulation environment was also used in prior works, see references in \cite{burghal2018band}.

The dataset has about $1150$ points, i.e., $|\mathcal{A}| = 1150$. The label that is associated with each point is whether the rate in the mmWave band is larger than the one in the cmWave band. To calculate the rate, we use the Shannon capacity with bandwidth and noise spectral density that are shown in Table \ref{Tab:SimEnvrionStc}.

%% file: Trajectories.tex
In this problem the dataset consists of {\em sequences} of features and labels that represent different UE trajectories and the designated BA decisions; each sequence can be viewed as an ordered subset of the available data points. The generation of such a dataset is challenging, as we have to generate correlated data points and reasonable trajectories. Note that the points on different trajectories may still be correlated as they belong to the same realization of the environment. As a result, we restrict our environment to one realization with several trajectories, i.e., we generate the trajectories over a grid that represents the cell. 

 We use a Semi Markov Smooth mobility model (SMS) to generate the motion trajectories \cite{zhao2006wsn03}. In an SMS model, the UE motion goes through cycles of four states until the end of the simulation time; it starts with the acceleration state with a random direction and a maximum ultimate speed, then a steady motion state, next it decelerates to zero before it stops in the last state, the UE can then go again to the first state. The duration of each state is a design parameter and can be set to random value. In our simulation, we omit the repeated data points (the consecutive points on the trajectories that correspond to the same location), and limit the number of repeated crossings over the same grid point. This model captures two important aspects of the realistic pedestrians' mobility, the smooth speed and direction adaptation (during the second state), and the possibility of changing the direction and speed along the route (at the beginning of the first state). 

%% file: LabelingDiagrams.tex
\begin{figure}
    
    \vspace{0mm}
\centering
 
\vspace{-0mm}
\centering
\includegraphics[width=1 \columnwidth, trim={1.5cm 0cm 0cm 0cm},clip]{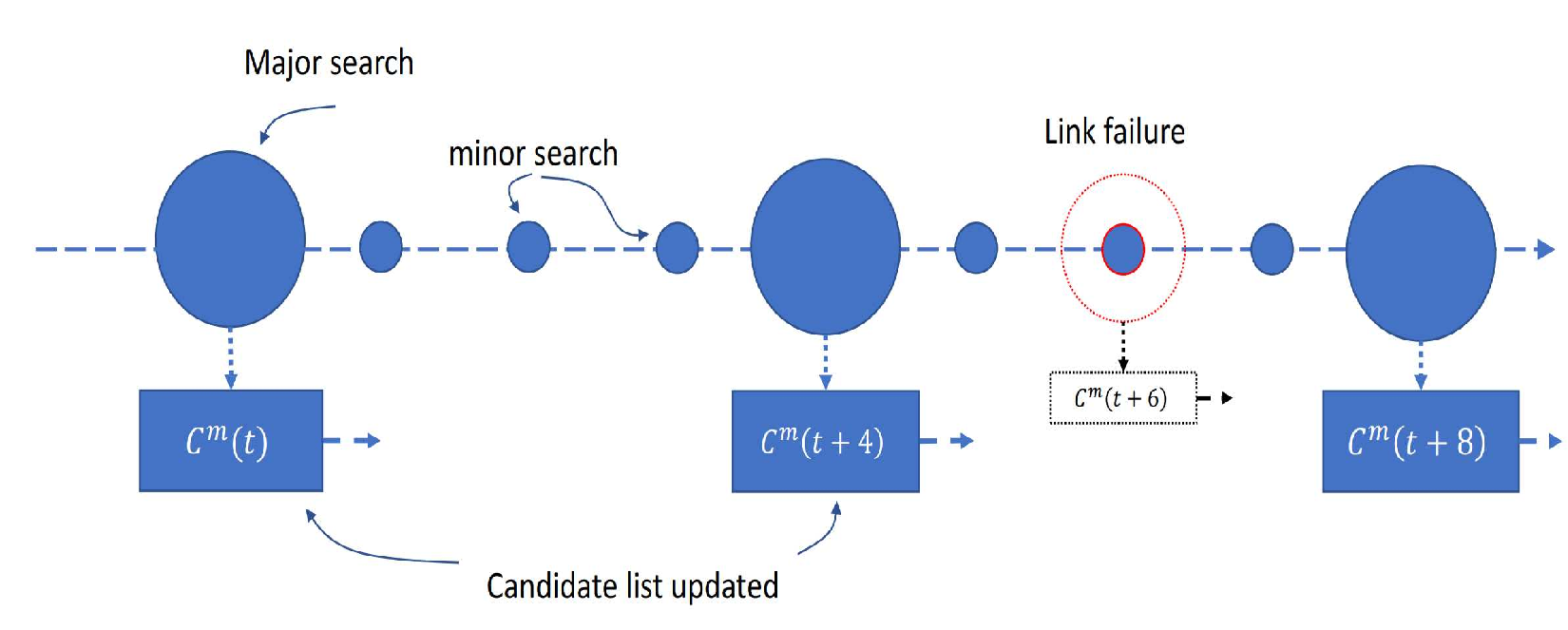}
 \vspace{-1 mm}
 \caption{A demonstration of the periodic major and minor beam searches and candidate lists in mmWave band with $N^m_p = 4$, note that $\mathcal{C}^m(t+1) = \mathcal{C}^m(t)$ after a minor search at {\it t}+1. Link failure could trigger a major search.}
\label{fig:mnrMajr}
   \vspace{-4mm}

\end{figure}

\begin{figure}

 
\vspace{-0mm}
  \centering
  \includegraphics[width=0.85\columnwidth, trim={0cm 0cm 0cm 0cm},clip]{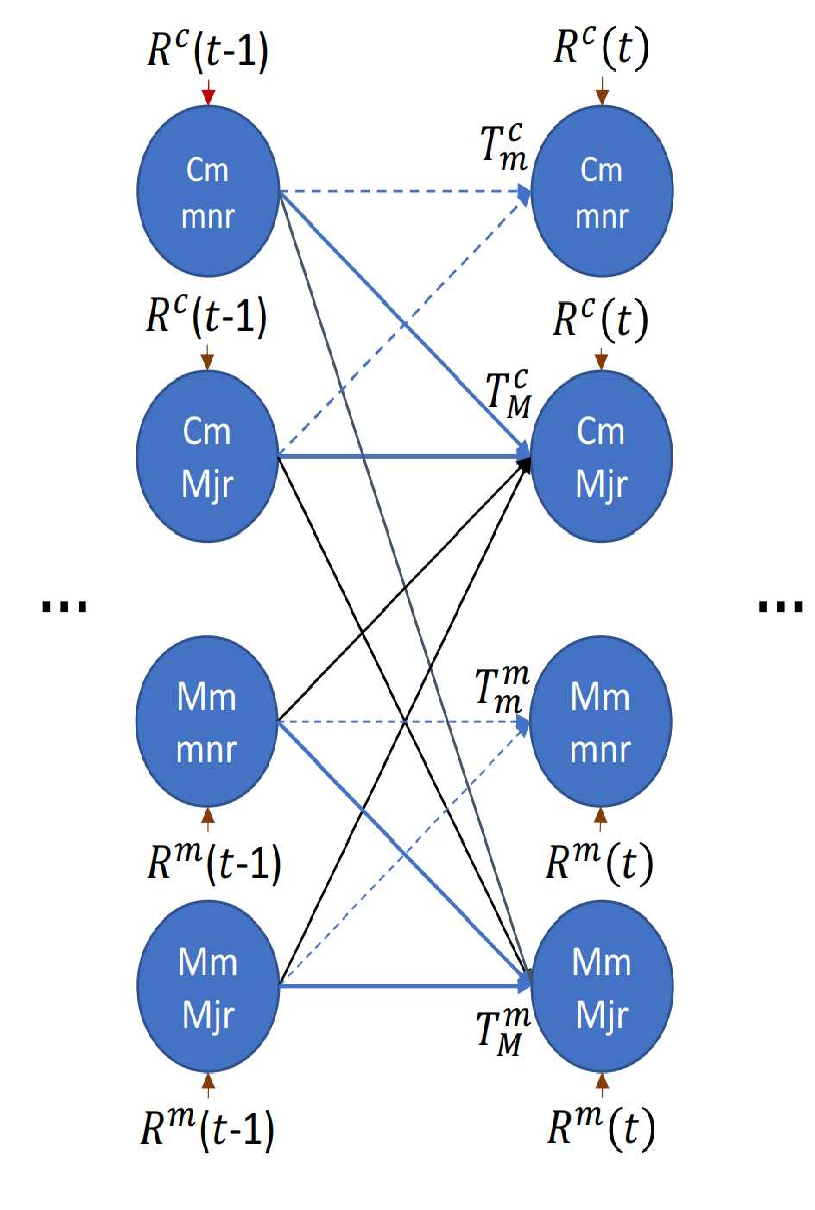} 
  \vspace{-6mm}
  \caption{The used four states (for two timeframes $t-1$ and $t$); Two states for minor and major searches in cmWave ("Cm mnr" and "Cm Mjr", respectively) and two states in mmWave ("Mm mnr" and "Mm Mjr"). Only the major search states can be connected to all previous states. The black lines indicate switching is possible from one band to another. Dashed lines indicate transitions within the same band. The rate $R^b(t)$ are calculated using eq. (\ref{eq:DPrate}).}

\vspace{-0mm}

\label{fig:VAstates}

    \vspace{-5mm}
\end{figure}

%% file: Preliminaries.tex
\subsubsection{Learning Techniques Overview}
Three ML solutions are considered in this paper: a solution that employs a recurrent NN (LSTM layers), multi-layer FC-NN only as another solution, and a simple GR (logistic regression) solution as a simple benchmark. Here we briefly introduce the three. The detailed architectures are in the following subsection. 
\begin{itemize}
    \item   GR uses a logistic function to map a linear combination of the input features to the range $[0,1]$, thus representing the probability of the output being in one of the classes. For instance, the output of GR (a soft decision) for an input with a single feature at time $\mathcal{F}(t)$, i.e., $|\mathcal{F}(t)|=1$, is $\tilde{\mathcal{D}}(t) = \mathbb{P}(\mathcal{F}(t)) = \frac{e^{\beta_0+\beta_1 \mathcal{F}(t)}}{1+e^{\beta_0+\beta_1 \mathcal{F}(t)}}$, where $\beta_0$ and $\beta_1$ are trainable parameters. 
    \item NNs consists of multiple layers each of which has a number of parallel neurons (nodes). The neuron performs a weighted combination of the input features and then passes it through a possibly non-linear transformation, also known as an activation function, e.g., a sigmoid function (generalization of the logistic function). Note that the GR can be viewed as a simple NN that consists of one neuron.
    \item LSTM is a popular recurrent NN architecture. The inputs of the LSTM are the current features, the previous output and the previous cell state. The cell state is a memory that is controlled by three gates, which control when to read, to write and to erase the value of the cell. The decisions of the gates are controlled by NNs that provide non-linear transformations of the input values.
\end{itemize} 

\begin{figure}
\vspace{-0mm}
\centering
\includegraphics[width=.5\columnwidth, trim={0cm 0cm 0cm 0cm},clip]{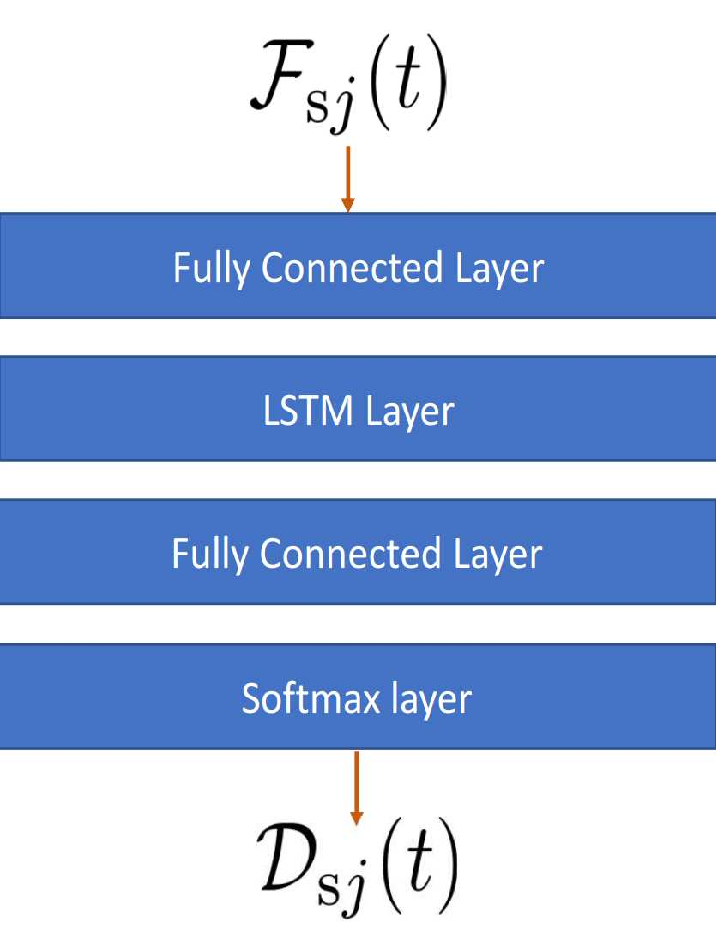} 
  \vspace{-1mm}
  \caption{The proposed LSTM based network structure.}

\vspace{-0 mm}
\label{fig:ProposedNW}

    \vspace{-9mm}
 
   \hfill
\end{figure}

The weights of the above solutions are determined during the training phase over a training dataset $\mathcal{A}^T$, where the goal is to minimize the prediction error. 

 \subsubsection{Training and Testing}
 
To train the learning approaches, we assume that the BS uses a dataset $\mathcal{A}^T = \{\mathcal{P}^T_1,...,\mathcal{P}^T_{N_T}\}$, where the {\em superscript} $T$ denotes training, and $N_T$ is the number of training examples. Each example sequence $\mathcal{P}^T_j$ consists of features-label pairs of sequences $(\mathcal{F}_{{\rm s}j},\mathcal{L}_{{\rm s}j})$, where $\mathcal{F}_{{\rm s}j}$ is the $j^{\rm th}$ sequence of features with length $T_j=|\mathcal{F}_{{\rm s}j}|$, and $F_{j} = |\mathcal{F}_{{\rm s}j}(t)|$ number of features. $\mathcal{L}_{{\rm s} j}(t)\in \{0,1\}$ is the $t^{\rm th}$ true label of the $j^{\rm th}$ sequence example. We assume that $\mathcal{A}^T$ is available to the BS, for instance through previous decisions or an initial network training phase. However, the procedure to acquire $\mathcal{A}^T$ is out of the scope of the paper.

For our study, let the set $\mathcal{A}$ denote the entire dataset in each environment, where each point $\mathcal{P}_j \in \mathcal{A}$ represents the features-label pairs of sequences with $N = |\mathcal{A}|$. In the simulation we \emph{randomly} split $\mathcal{A}$ into mutually exclusive sets: training set $\mathcal{A}^T$ and a testing set $\mathcal{A}^S$. We may further split $\mathcal{A}^T$ into a training subset $\mathcal{A}_{\rm t}^T$ and a validation subset $\mathcal{A}_{\rm v}^T$. The sizes of $\mathcal{A}_{\rm t}^T$,$\mathcal{A}_{\rm v}^T$ and $\mathcal{A}^S$ are respectively $N_{T},N_{V},$ and $N_{S}$. More details about this are provided in the next sections. During the \emph{training phase}, as it is commonly used in binary classification problems, we employ the Cross Entropy (CE) cost function, i.e., $$\bar{\mathcal{E}}_{{\rm CE},X} = \frac{1}{N_X}\sum_i^{N_X} (-\mathcal{L}_i log(\tilde{\mathcal{D}}_i)-(1-\mathcal{L}_i)log(1-\tilde{\mathcal{D}}_i)).$$ 
As in Sec. \ref{subsec:DLPerfMet}, we drop the indices $j$ and $t$, and use the subscript $X$ to distinguish the used dataset to calculate these values. For the introduced datasets we have $X\in\{T,V,S\}$. 

\subsection{Sequential BA}
The proposed network architecture is shown in Fig. \ref{fig:ProposedNW}. It consists of an FC layer with $20$ neurons, LSTM layer of size $40$, followed by another FC layer with $20$ neurons and then a softmax layer. We noticed an improvement in the performance when we introduced the two FC layers, as they might help scale the values of the input and the output of the network. Since the proposed network is relatively simple, we adopt ensemble learning, where we train five networks using a subset of the training data, and report the average result. We use the {\em Adam} algorithm for training \cite{kingma2014adam}. The choice of the above architecture and other hyper-parameters is done using cross validation.

As a simple ML benchmark, we use GR-based (denoted by $\rm GR_{H}$), and provide an NN-based as another solution (denoted by $\rm NN_{H}$). For $\rm NN_{H}$ we use two hidden layers with $70$ and $40$ neurons, respectively. For these solutions, and since the decisions depend on the previous data points, we use a modified dataset ${A^{T}}'$. In particular, to utilize the $Q$ previously observed points in $\rm GR_{H}$ and $\rm NN_{H}$, their datasets should have input features size $|\mathcal{F}'_{{\rm s}j}(t)| = (Q+1)F_{{\rm s}j}$, i.e., the size is equal to the number of used features in the current and the $Q$ previous points.

%
%

%% file: GP_Based_BA.tex
Based on the model introduced in Sec. \ref{subsec:GPsolModl}, we denote the following two events
\begin{align*}
\mathcal{W}_b(t) : R^b(t) > R^{b'}(t), ~~ b \neq b', b,b' \in\{c,m\}.
\end{align*} 
Then given the set of observations at time frame $t$ $\mathcal{H}_t$, a reasonable band assignment rule is 
\begin{align}\label{eq:Dprule}
b^* = \arg\max_{b \in \{c,m\}}~~ \mathbb{P}(\mathcal{W}_b(t)|\mathcal{H}_t).
\end{align}
This is a Maximum A posteriori Probability (MAP) Decision rule \cite{kay1998fundamentals}, which has the following property.
\begin{ppro}
Minimum BA error probability can be achieved when the BS chooses band $b^*$ that satisfies 
$$b^*: \mathbb{P}(\mathcal{W}_{b^*}(t)| \mathcal{H}_t)\geq 0.5.$$
\end{ppro}
\begin{proof}
The proof is simple. From (\ref{eq:Dprule}) $b$ is chosen when $ \mathbb{P}(\mathcal{W}_b(t)|\mathcal{H}_t) \geq  1- \mathbb{P}(\mathcal{W}_{b}(t)|\mathcal{H}_t)$.
Next, by the definition of the MAP rule, the BS minimizes the probability of error at \emph{each} time frame, which as a result minimizes the overall probability of BA error.
\end{proof}
The proposition indicates that the natural choice for $\gamma_T$ is optimal under the conditions above.
 \subsubsection{Exact Solution}
For a sequential BA problem we have $\mathcal{H}_t = \{S^c(t-Q),S^c(t-Q+1),...,S^c(t),S^m(t-Q),S^m(t-Q+1),...,S^m(t) \}$. The goal is to choose either $\mathcal{W}_c(t+U)$ or $\mathcal{W}_m(t+U)$. Thus we choose $b^* = m$ if:
\begin{align}\label{eq:DpGGPdcRule}
  \mathbb{P}(\mathcal{W}_m(t+U)|\mathcal{H}_t) \geq 0.5,
\end{align}
and $b^* = c$ otherwise. Since the set $\mathcal{H}_t$ has more than a single value, we need to use the joint Gaussian assumption over time (space). Then (\ref{eq:DpGGPdcRule}) becomes
\begin{align}\label{eq:DpCondRule}
 & \mathbb{P}(\mathcal{W}_m(t+U)|\mathcal{H}_t)= \mathbb{P}(R^m(t+U)\geq R^c(t+U)|\mathcal{H}_t)=  \nonumber \\ & \int_{-\infty}^{\infty} \mathbb{P}( R^m(t+U) \geq R^c(t+U) | \mathcal{H}_t, S^c(t+U) = s) \nonumber \\ & ~~~~~~ f_{S^c| \mathcal{H}_t}(s)ds,
\end{align}
where $f_{S^c|\mathcal{H}_t}(s)$ is the probability density function (PDF) of $S^c(t+U)$ given $\mathcal{H}_t$, which follows a normal distribution with mean $\mu_{c|H}$ and variance $\sigma^2_{c|H}$. Note that we conditioned on $S^c(t+U)$ and used the integration to circumvent the fact that the probability of $\mathcal{W}^m : \{ R^m> R^c\}= \{ R^m- R^c>0\}$ cannot be calculated using a simple probability distribution without some approximations (as we discuss in the next subsection). Next, using $(\ref{eq:rate})$ we rewrite (\ref{eq:DpCondRule}) as:
\begin{align}\label{eq:DpIntgSimpl}
& \int_{-\infty}^{\infty} \mathbb{P}( S^m(t+U) \geq \mathcal{V}_2(s) | \mathcal{H}_t, S^c(t+U) = s) f_{S^c| \mathcal{H}_t}(s)ds, \\ \nonumber
 & {\rm where,~} \\ \nonumber & \mathcal{V}_2(s) = \frac{1}{\gamma} \log_{10}\big(\frac{1}{\gamma_m'}\big(\exp\big(\frac{\omega_c}{\omega_m}\log(1+\gamma_c' 10^{\gamma s})\big)-1\big)\big). \nonumber
\end{align}
To evaluate (\ref{eq:DpIntgSimpl}), we first point out that $S^m(t+U)$, $S^c(t+U)$, and the observations in $\mathcal{H}_t$ are jointly normal, thus it is enough to calculate the conditional mean and variance of $S^m(t+U)$ given $\mathcal{H}_t$ and $S^c(t+U)$. The details are presented in the appendix-A.

\subsubsection{Approximate Solution}
We can provide a simpler rule that does not rely on integration by assuming that $w_b\log(1+10^{{\rm SNR}^b(t)})\approx w_b\log(10^{{\rm SNR}^b(t)}) $, which is usually referred to as the "high SNR assumption". Then we have:
\begin{align*}
\tilde{R}^b = \omega_b \log(\gamma_b) + \frac{\log(10)}{10} \omega_b S^b,
\end{align*}
which follows a normal distribution with mean and variance, respectively,
$\tilde{\mu}_b$ and $\tilde{\sigma}^2_{b}$. 
Then we can define the event $\tilde{\mathcal{W}}_b(t+U) : \tilde{R}^b (t+U)\geq \tilde{ R }^{b'}(t+U)$, and choose $b$ that satisfies $\mathbb{P}(\tilde{\mathcal{W}}_b| \mathcal{H}_t)> 0.5$. To calculate this probability, and taking $b= m$ and $b'= c$, we have:
$\mathbb{P}(\tilde{\mathcal{W}}_m| \mathcal{H}_t)= \mathbb{P}(\tilde{ R }_D\geq 0| \mathcal{H}_t)$. 
where $\tilde{ R }_D = \tilde{ R }^m (t+U)-\tilde{ R }^c(t+U)$, since $S^m$ and $S^c$ are jointly normal, so are $\tilde{ R }^m$ and $\tilde{ R }^c$, and thus $\tilde{ R }_D$ is normally distributed with mean and variance, $\mu_D$ and $\sigma_{D}^2$. The decision rule can be shown to be 
\begin{align}\label{eq:GPApprule}
\tilde{\mu}_c-\tilde{\mu}_m \leq \Sigma_{D,H} \Sigma_{H}^{-1} \mathbf{h},
\end{align}
where $\Sigma_{D,H}$ captures the join covariance of $\tilde{ R }_D$ and the observations. Due to the simplicity of this rule one can easily derive a number of interesting quantities, such as the probability of error, however, we omit such discussion due to space limitations. We hereafter refer to the exact and approximate GP based solutions, respectively, as $\rm GP$ and $\rm GP_{App}$. In Sec. \ref{Sec:StochEnvDL}, we study the impact of the decision threshold $\gamma_T$ and observation window $Q$ on both solutions.
%
%
%

%% file: Stoch_Seq.tex
\subsubsection{Data Generation}
The main parameters were introduced in Sec. \ref{sec:stochData}. We assume the BS is located at the center of a square cell with a side length of $500$ m, with a $5$ m separation distance between the points on the grid. We generated $1000$ sequences, we assumed that the duration for each sequence is $900$ s with {\em speed up to} $1.5$ m/s and a $4$ s sampling period.\footnote{ With the values of $d_{\rm dcor}$ in Table \ref{Tab:SimEnvrionStc}, the sampling period provides enough shadowing samples based on Nyquist theorem.} To generate the shadowing values we use the correlation model in (\ref{eq:xcorr}) for two different $\nu$ values, $\nu = 1$ and $\nu = 1.9$. We assume that the observation window of GP-based and learning-based solutions (other than LSTM-based) is $Q = 5$. We use $70\%$ of the sequences for training and $30\%$ for testing. For the LSTM based solution we use $50$ sequences for cross validation.
\subsubsection{Performance ($\nu = 1$)}
\input{Tables/Nu1andNu19}
The results are presented in Table \ref{Tab:ResultsNu1andNu19}. The first five rows show the feature combinations. We notice that $\rm GP_{App}$ shows around $9\%$ performance degradation compared to $\rm GP$. For learning schemes, we notice that all-features combination (c-5) provides the best performance followed by the combination of cmWave power and location (c-4). One reason for (c-5)'s good performance seems to be the location information. This conjecture is supported by the performance of (c-1) compared to having cmWave and mmWave powers (c-6). The importance of location information is intuitive, as it relates to trajectory prediction which in turn impacts the channel conditions. The performance difference between the LSTM-based solutions and $\rm NN_{H}$ may be attributed to the inherent ability of the LSTM layer for sequential learning. Note that the cmWave and mmWave power combination (c-6) still provides valuable information, and with it as the basis, most of the learning schemes outperform $\rm GP$, and all of them outperform $\rm GP_{App}$. This is important as both approaches, the GP-based and the ML-based, may use cmWave and mmWave power as input observations. Interestingly, we also observe that using a cmWave only (c-7) learning scheme can outperform $\rm GP_{App}$. Fig. \ref{Fig:PeVsU} shows the performance as a function of $U$ for (c-5) and (c-6). It shows that the performance of the proposed LSTM-based solution dominates the other schemes for small $U$, but the probability of BA error increases logarithmically as $U$ increases. 

The use of $\gamma_{T} = 0.5$ for $\rm GP$ was justified in Sec. \ref{sec:GPBasedSol}. In Fig. \ref{Fig:PeVsGamma} we present the impact of $\gamma_T$ on the performance for combinations (c-5) and (c-6), and show the performance for the $\rm LSTM$ (as listed in the table) for comparison. We notice that $\gamma_T \approx 0.5$ is good for most of the schemes except $\rm GP_{App}$, indicating that $\rm GP_{App}$ can be improved with a judicious choice of $\gamma_T$.

\input{UandGamma}

%
\subsubsection{Performance ($\nu = 1.9$)}\label{subsec:DPNu1.9}
With $\nu = 1.9$, the correlation function decays faster than above, however, we noticed that the impact of prior observations is more pronounced. The results for several features observations are presented in Table \ref{Tab:ResultsNu1andNu19}. For GP-based, we observe that $\rm GP$ outperforms $\rm GP_{App}$ by about $20\%$, and several learning schemes outperform the $\rm GP$ with several features combinations. In addition, the all-features combination (c-5) still has the least $\bar{\mathcal{E}}_S$, but different than above the performance gain is attributed to the power in the two bands (c-6). Interestingly, the LSTM-based solution using cmWave power (c-7) is as good as location (comparable to $\rm GP_{App}$) in this environment, which indicates that (c-7) is a good BA predictor. The significance of (c-7) is also evident for other learning schemes that use $Q$ observations. 


Based on the used pedestrians' speed, grid points separation and sampling interval we anticipate that observations outside the used observation window, of size $Q=5$, have small influence on the BA at time frame $t+U$. However, considering the adopted motion model, this may not be accurate, as an old observation might be highly correlated (closely located) to future value. This complicates the analysis of the impact of $Q$. Instead we here restrict our attention to a simple circular motion around the BS, where we consider $5000$ sequences, each corresponding to one circle around the BS and having an independent shadowing realization; we here relax some of the correlation assumptions since we consider only cmWave and mmWave power information. The results are provided in Fig. \ref{Fig:Nu19CirclevsQ}. The learning schemes achieve the same performance compared to the {\em optimal} solution ($\rm GP$ in this case). Starting with $\rm GP$ vs. $\rm GP_{App}$, it is clear that the approximation introduces an error floor for $\rm GP_{App}$. The $\rm GP_{App}$ shows a noticeable decrease in $\bar{\mathcal{E}}_S$ until $Q=4$, as it may reduce the uncertainty, however beyond that the error increases again due to the model mismatch. The slow improvement for large $Q$, could be attributed to the decrease of added information in older observations in such a uniform motion. A more comprehensive study is needed for this problem but it highly depends on the motion model.

\subsubsection{Additional Remarks} Two important remarks about the solutions.\\
{\em Remark 1:} The time or the number of operations for inference might be important for some real time systems. Table \ref{Tab:AvgInfTime} shows an example of the inference times based on our (not optimized) MATLAB implementation on a personal computer that uses Intel i7 processor with $8$ GB RAM.  
   \begin{table}[!htb]
 \vspace{2mm}
    \centering 
 \begin{tabular}{|c|c|c|c|c|c|}
 \hline
\bf Solution & \bf LSTM based& \bf NN\textsubscript{H} & \bf GR\textsubscript{H}   & \bf   \bf  GP  &  \bf GP\textsubscript{App}       \\
\hline
 \bf Average Inference Time & 300 & 201 &   0.44 & 1700 & 214 \\ \hline
 \end{tabular}
 \vspace{2mm}
  \caption{Average inference time in $\mu $s. The LSTM solution assumes parallel implementation for ensemble inference. Here dual power features are used.}
   \vspace{2mm}
 \label{Tab:AvgInfTime}
\end{table}
\\
{\em Remark 2:} Although the performance of the GR solution can be improved, e.g., with a non-linear feature mapping, that is out of the scope of this work, as the goal here is to use it as a simple ML benchmark.

%% file: Tables/Nu1andNu19.tex
\begin{table*}[!htb]
\vspace{1mm}
    \centering
 \begin{tabular}{|c|c|c|c|c|c|c|c|c|}
 \hline

  \bf Combination   &                  		& {\bf c-1 } & \bf c-2   & \bf c-3   &\bf c-4    & \bf c-5    & \bf c-6   & \bf c-7    \\
\hline
\multirow{4}{*}{\bf Feature}& $\boldsymbol{d}$                           		 &\checkmark  &\checkmark &           &\checkmark &\checkmark  &           &            \\
 \cline{2-9}
 
{}&  $\boldsymbol{\theta}$                      		&\checkmark  &           &\checkmark &\checkmark &\checkmark  &           &             \\
 \cline{2-9}
{}& \bf cmWave       Power                 		&            &\checkmark &\checkmark &\checkmark &\checkmark  &\checkmark & \checkmark \\
 \cline{2-9}
{}& \bf mmWave    Power                   		&            &           &           &           &\checkmark  &\checkmark &\\

  \hline 
  
  \hline  
  
  \multirow{5}{*}{\bf Solution}& \bf  LSTM based   	&  \bf{.178}/\bf{.166}  & \bf{.193}/\bf{.158}	&\bf{.183}/\bf{.142}	 & \bf{.165}/\bf{.123} &\bf{.163}/\bf{.107}	 &\bf{.191}/\bf{.119}	 &\bf{.206}/\bf{.166}   \\
 \cline{2-9}
 {}& \bf NN\textsubscript{H}    		&  .211/.196 & .216/.18 &.197/.155	 & .192/.147  &.189/.118	 &.209/.136 &.22/.176   \\
 \cline{2-9}
 {}& \bf GR\textsubscript{H}  		&  .22/.21  & .22/.18	&.214/.169 	 & .209/.172 &.198/.127 	 &.211/.133	 &.221/.179   \\
 \cline{2-9}  
 &  \bf GP     &\multicolumn{7}{c|}{ .201/.126}  \\
   \cline{2-9}
 &    \bf GP\textsubscript{App}    &\multicolumn{7}{c|}{.221/.159}  \\
 
  \hline 
  
 \end{tabular}
 \vspace{-0mm}
\caption{The misclassification error (${\bar{\mathcal{E}}}_S$) in the stochastic data under different feature availability, results for $\nu =1 $ and $ \nu =1.9$, for each case the error values are separated by $"/"$. The prediction target is at $U=4$. On average $47.9$\%  and $48.4$\% of the labels are $"1"$ when $\nu =1 $ and $ \nu =1.9$, respectively. }
   \vspace{-3mm}
 \label{Tab:ResultsNu1andNu19}
\end{table*}
 

%% file: UandGamma.tex
\begin{figure}[htb]    
\vspace{0mm}
\centering
\vspace{-0mm}
\includegraphics[width=0.9\columnwidth, trim={0.0cm 0cm 0cm 0cm},clip]{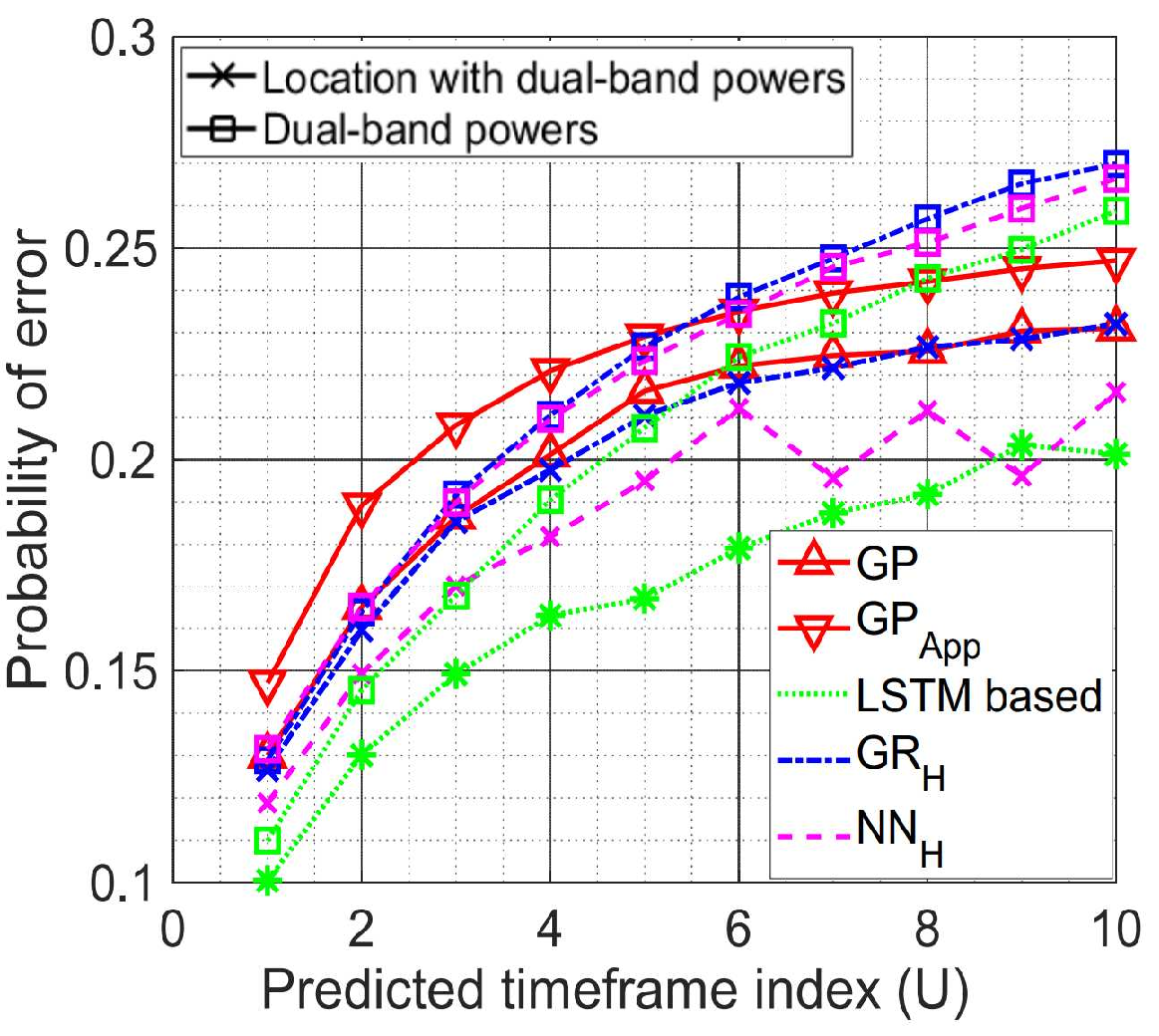}
\vspace{-1mm}
\caption{Misclassification error $\bar{\mathcal{E}_S}$ vs. $U$  for two combination: (c-5) and (c-6) in Table \ref{Tab:ResultsNu1andNu19} for in $\nu = 1$.}
\label{Fig:PeVsU}
 \vspace{-4mm}
 \end{figure}
 
 \begin{figure}[htb]    
\vspace{-0mm}
\centering
\vspace{-0mm}
\includegraphics[width=0.9\columnwidth, trim={0.1cm 0cm 0cm 0.0cm},clip]{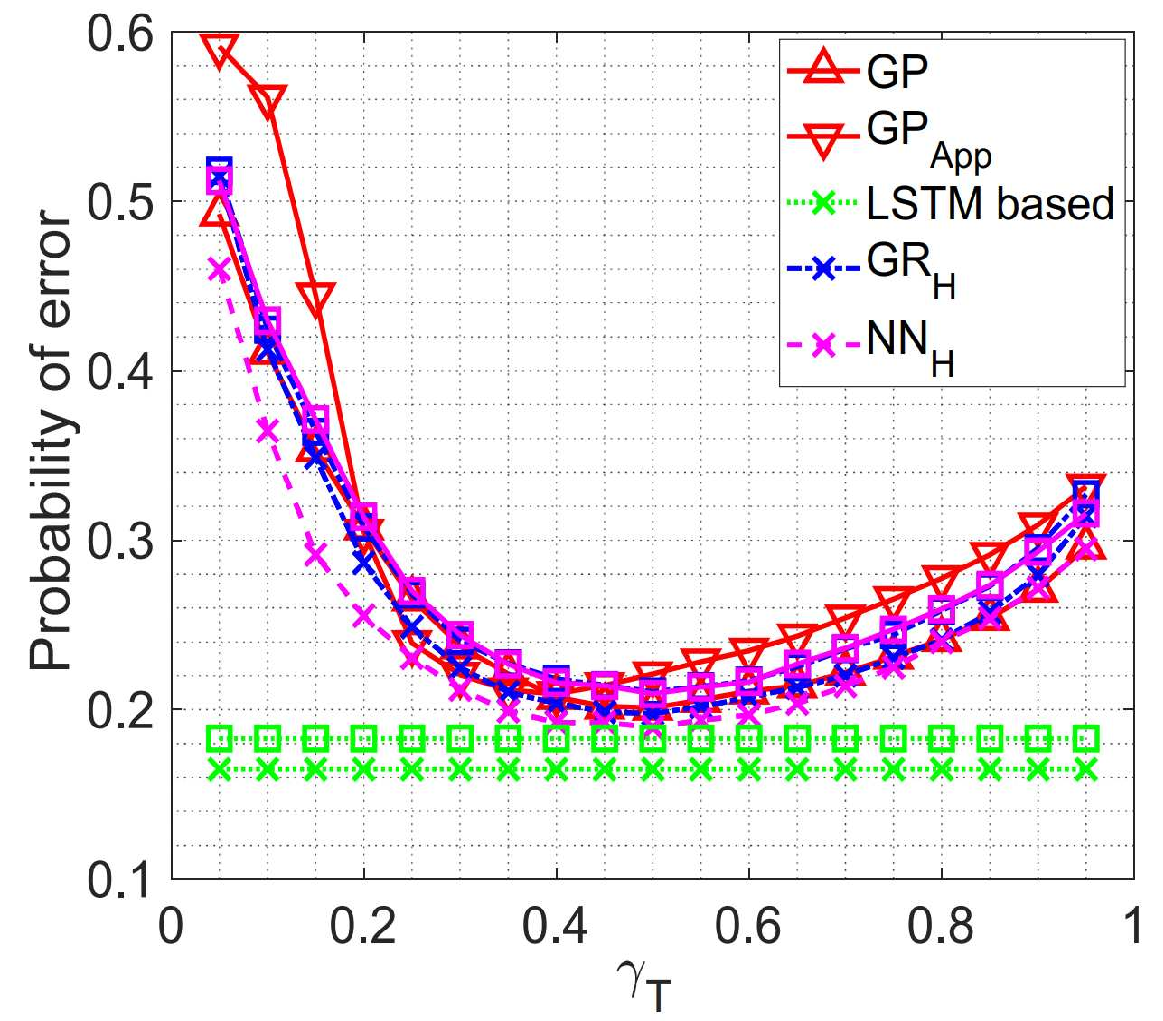}
\vspace{-1mm}
\caption{Misclassification error $\bar{\mathcal{E}_S}$ vs. the decision threshold $\gamma_T$ with for $\nu =1$ for features combinations: (c-5) and (c-6) in Table \ref{Tab:ResultsNu1andNu19}.}
\label{Fig:PeVsGamma}
 \vspace{-7mm}
 \hfill
\end{figure}

%% file: RT_Seq.tex
\subsubsection{Data Points}
To generate the sequences we use the motion model discussed in Sec. \ref{sec:genSeq} over the ray-tracing grid. This includes the pedestrians speed, simulation time, and sampling interval. We generate $1000$ sequences, and use $350$ of them for training out of which $70$ are chosen randomly for cross validation for the LSTM-based solution. For the GP-based solutions, we extract the channel parameters to fit the path-loss (using a double linear fit) and then compute the parameter for the correlation model (\ref{eq:xcorr}) and (\ref{eq:Dpcov}).

\subsubsection{Performance}

\input{QandRTU} 
\input{Tables/RTU4}
Table \ref{Tab:SeqResultsRTU4} shows the performance of the schemes in this environment with a structure similar to the one in Sec. \ref{Sec:StochEnvDL}. We start by observing that both GP-based solutions are not performing well. This can be explained by our investigation of the shadowing distribution in this environment, where we observed that it is far from satisfying the GP assumption even for a single band. We also note that the $\rm GP$ has a larger $\bar{\mathcal{E}}_S$ compared to $\rm GP_{App}$ ($14\%$ worse for $U=4$); this surprising result might be related to the fact that $\rm GP$ is only exact if the Gaussian model is fulfilled, so that an approximate algorithm might suffer less in the presence of a model mismatch. For the learning schemes, we focus on the performance of LSTM-based and $\rm NN_{H}$ solutions. We notice that the location plus the powers in both bands (c-3) still achieves low $\bar{\mathcal{E}}_S$, however, note that for the LSTM-based solutions this is the case for other combinations as well such as the powers in both bands plus AoA and Delay (c-9). Comparing $\bar{\mathcal{E}}_S$ for the power in both bands (c-4), location (c-1), and AoA and Delay (c-7), we notice that (c-4) plays a major role in the performance gain that we observed. The value that (c-4) achieves is also possible using other combinations, namely (c-8) and (c-10), which require the cmWave power plus other features, indicating the practicality of the solutions when only the cmWave power (e.g., through a control signal) is periodically observed. Note that, for the LSTM-based scheme, $\bar{\mathcal{E}}_S$ for (c-1) is not much worse than (c-4), which explains why the {\em combination} of location and cmWave power (c-2) would be as good as (c-3). For $\rm NN_{H}$, things are slightly different as the observed performance gain is mainly attributed to the location information (c-1), which alone provides a performance comparable to (c-9).
   
    

The impact of $U$ is shown in Fig. \ref{Fig:PeVsURT}. The probabilities of error $\bar{\mathcal{E}}_S$ using (c-3) and (c-9) are comparable over different $U$, which is interesting as this could eliminate the need for explicit feedback of the location information. For the GP based schemes, as discussed above, $\rm GP_{App}$ is better than $\rm GP$. One reason for the relatively better performance can be attributed to the intuitive $\rm GP_{App}$ structure, eq. (\ref{eq:GPApprule}), which is a threshold rule that employs the gap between the average received powers in the two bands, which may rely less on the impact of the GP assumption on the rates. Note that the general behavior of the GP-based solution can be explained by the fact that the environment does not follow the GP assumption anymore, but rigorous explanations are difficult since we here have a single environment realization. Finally, we discuss the impact of feature uncertainty on the performance in appendix-B.

%% file: QandRTU.tex
\begin{figure}
 
    
\vspace{0mm}
\centering
\vspace{-0mm}
\includegraphics[width=0.9\columnwidth, trim={0.2cm 0cm 0cm 0.0cm},clip]{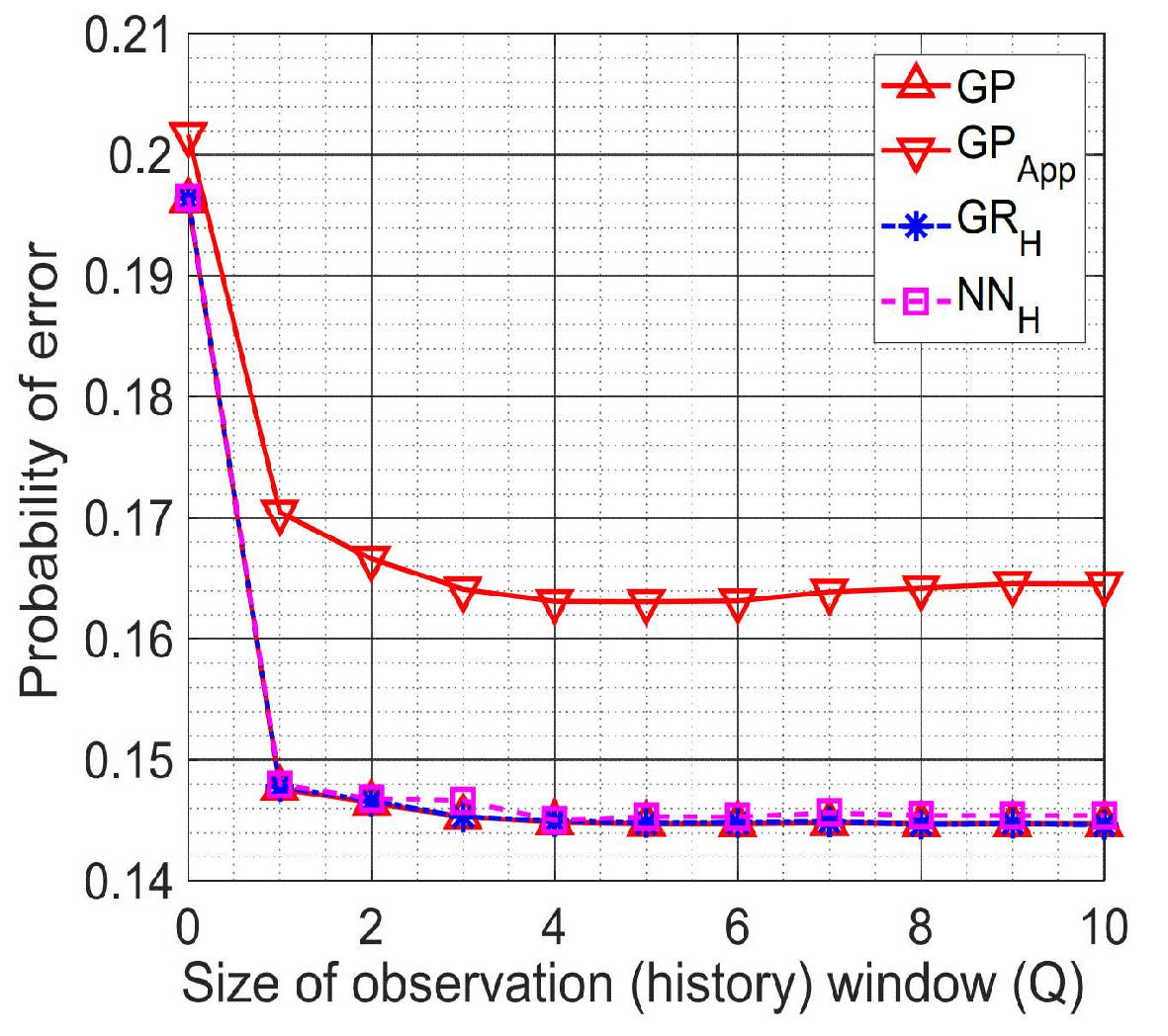}
\vspace{-2mm}
\caption{The impact of the size of the observation window $Q$ in stochastic environment with circular trajectory. For this data set the percentage of "1"s is $75\%$.}
\label{Fig:Nu19CirclevsQ}
 \vspace{-0mm}
 \end{figure}
 
 \begin{figure}[htb]
\vspace{-0mm}
\centering
\vspace{-0mm}
\includegraphics[width=0.9\columnwidth, trim={0cm 0cm 0cm 0.0cm},clip]{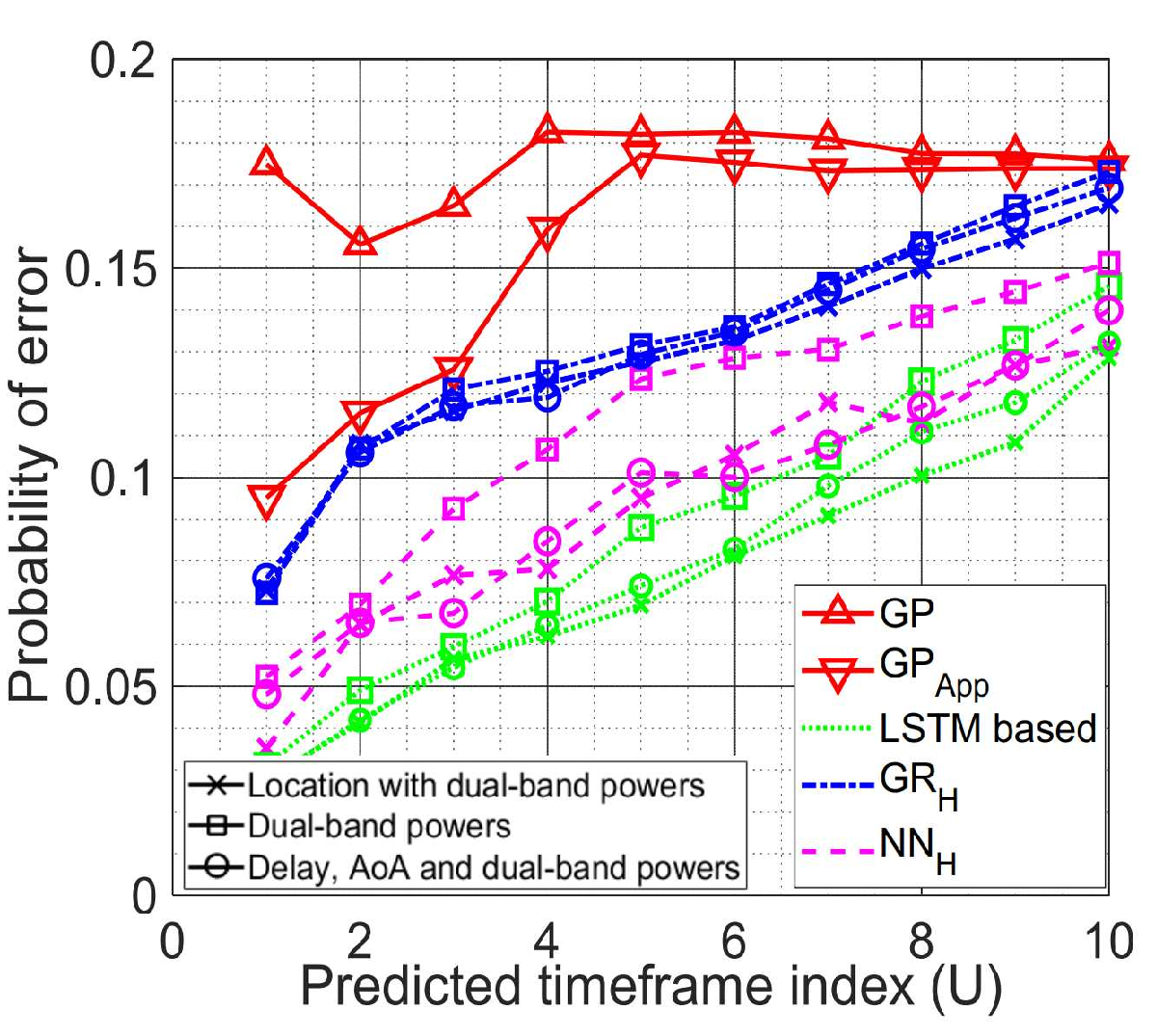} 
\vspace{-1mm}
\caption{BA misclassification error $\bar{\mathcal{E}_S}$ vs. $U$ for three features combinations (c-3), (c-4) and (c-9) in Table \ref{Tab:SeqResultsRTU4}.}
\label{Fig:PeVsURT}
 \vspace{-10mm}
 \hfill
\end{figure}

%% file: Tables/RTU4.tex
\begin{table*}[!htb]
\vspace{2mm}
\centering
\begin{tabular}{|c|c|c|c|c|c|c|c|c|c|c|c|c|}
\hline
\bf Combination  &   	  & \bf c-1   & \bf c-2   &\bf  c-3   &\bf c-4     & \bf c-5   & \bf c-6   	&\bf c-7 	& \bf c-8 	& \bf c-9 		& \bf c-10 	& \bf c-11  \\
\hline	
 \multirow{6}{*}{\bf Feature} &   $\boldsymbol{d}$      &\checkmark &\checkmark &\checkmark  &           &           &			&           &           &           	&          	&           \\
\cline{2-13}
& $\boldsymbol{\theta}$ &\checkmark &\checkmark &\checkmark  &           &           &           	&           &           &           	&\checkmark &          \\
\cline{2-13}
& \bf cmWave            &           &\checkmark &\checkmark &\checkmark  &\checkmark &			&			&\checkmark & \checkmark	&\checkmark &\checkmark\\
\cline{2-13}
& \bf mmWave            &           &           & \checkmark &\checkmark &		   & \checkmark &		  	&           &  \checkmark	&          	&          \\
\cline{2-13}
& \bf Delay             &           &           &           &			   &           &           	&\checkmark &\checkmark &\checkmark		&          	& \checkmark\\
\cline{2-13}
& \bf AoA               &           &           &           &            &           &           	&\checkmark	&\checkmark &\checkmark		&			&          \\
\hline 
 \multirow{5}{*}{\bf Solution}&\bf LSTM based  	 &\bf{ .077}   &\bf{.064}   &\bf{.062}    &\bf{.07}    &\bf{.094}   &\bf{.08}     &\bf{.096}  &\bf{.073}   &\bf{.065}       &\bf{.071} 	&\bf{.079} \\
\cline{2-13}
&\bf NN\textsubscript{H}	      	  &.086   &.083   &.076     &.107    &.134   &.12     &.113   &.098   &.085     &.099 	&.107 \\

\cline{2-13}
&\bf GR\textsubscript{H}       	  &.208   &.134   &.123  &.125    &.142  &.122     &.224   &.138   &.119       &.137 	&.14 \\

\cline{2-13}
&\bf GP  	 &\multicolumn{11}{c|}{.183  }  \\
\cline{2-13}
&\bf GP\textsubscript{App}   	 &\multicolumn{11}{c|}{.159 }  \\

\hline
\end{tabular}
\vspace{-0mm}
\caption{BA misclassification error (${\bar{\mathcal{E}}}_S$) in the ray-tracing environment, for SISO system under different feature availability and $U=4$. The percentage of points with labels equal to $"1"$ is approximately $30$\%.}
\vspace{-4mm}
\label{Tab:SeqResultsRTU4}
\vspace{-0mm}
\end{table*}

%% file: Tables/TblMIMOSeq.tex
\begin{table*}[!htb]
\vspace{2mm}
\centering
\begin{tabular}{|c|c|c|c|c|c|c|c|c|c|c|c|c|}
\hline
\bf Combination &  	  & \bf c-1   & \bf c-2   &\bf  c-3   &\bf c-4     & \bf c-5   & \bf c-6   	&\bf c-7 	& \bf c-8 	& \bf c-9 		& \bf c-10 	& \bf c-11  \\
\hline	
  \multirow{6}{*}{\bf Feature}  & $\boldsymbol{d}$      &\checkmark &\checkmark &\checkmark  &           &           &			&           &           &           	&          	&           \\
\cline{2-13}
&$\boldsymbol{\theta}$ &\checkmark &\checkmark &\checkmark  &           &           &           	&           &           &           	&\checkmark &          \\
\cline{2-13}
&\bf cmWave            &           &\checkmark &\checkmark &\checkmark  &\checkmark &			&			&\checkmark & \checkmark	&\checkmark &\checkmark\\
\cline{2-13}
&\bf mmWave            &           &           & \checkmark &\checkmark &		   & \checkmark &		  	&           &  \checkmark	&          	&          \\
\cline{2-13}
&\bf Delay             &           &           &           &			   &           &           	&\checkmark &\checkmark &\checkmark		&          	& \checkmark\\
\cline{2-13}
&\bf AoA               &           &           &           &            &           &           	&\checkmark	&\checkmark &\checkmark		&			&          \\
\hline  
  \multirow{6}{*}{\bf Solution}  & \multirow{2}{*}{\bf LSTM based} ~  ($\boldsymbol{\bar{\mathcal{E}}_S}$)	  &  \bf{.091}  &   \bf{.084} &   \bf{.079}   &  \bf{.089}  &  \bf{.119} &  \bf{.097} &  \bf{.115} &  \bf{.088} &  \bf{.081}   &  \bf{.094}	& \bf{.1}\\
&  ~~~~~~~~~~~~~~~~~~~ ($\boldsymbol{\bar{\mathcal{R}}_S}$)	 &  \bf{.056} &  \bf{.055}  &   \bf{.052}  &   \bf{.064} &  \bf{.083} &   \bf{.072}   &   \bf{.072} &  \bf{.055}  &   \bf{.055}  &  \bf{.061}	&  \bf{.067} \\
\cline{2-13}
& ~~~~~\multirow{2}{*}{\bf NN\textsubscript{H}} 	    ~~~~~~~ ($\boldsymbol{\bar{\mathcal{E}}_S}$)	  &  .093 & .093 &  .087   &  .095 & .14  &  .125   & .152 & .097 &   0.088  & .102	&  .107\\
&  ~~~~~~~~~~~~~~~~~~~ 	     ($\boldsymbol{\bar{\mathcal{R}}_S}$)   &  .058  &   .072  &   .065  &  .076 &  .11  &  .122   &  .114  &  .071  &  .064  &  .069	&  .077 \\

\cline{2-13}
& ~~~~~\multirow{2}{*}{\bf GR\textsubscript{H}} 	    ~~~~~~~ ($\boldsymbol{\bar{\mathcal{E}}_S}$)	  & .221  & .147 &  .135  &    .14 & .158  & .137 & .242 & .155  & .141    & .149	&  .156\\
&  ~~~~~~~~~~~~~~~~~~~ 	      ($\boldsymbol{\bar{\mathcal{R}}_S}$)  & .208  & .111 & .113    &.118 & .122 &  .119   & .208  & .12 & .119     & .111	& .121\\

\hline 
\end{tabular}
\vspace{-0mm}
\caption{Performance (the misclassification error (${\bar{\mathcal{E}}_S}$) and the normalized rate loss (${\bar{\mathcal{R}}_S}$)) of the solutions on ray-tracing data for MIMO system under different feature availability.}
\vspace{-4mm}
\label{Tab:MIMOSeqResultsRT}
\vspace{-0mm}
\end{table*}